\documentclass[journal]{IEEEtran}
\usepackage[T1]{fontenc}

\usepackage{cite}
\usepackage{amsmath}
\usepackage{amsthm}
\usepackage{amssymb}
\usepackage{dsfont}
\usepackage{graphicx}
\usepackage{multicol}
\usepackage{textcomp}
\usepackage{xcolor}
\usepackage{xspace}
\usepackage{bbm}
\usepackage{algorithm,algorithmic}
\usepackage{hyperref}
\usepackage{tabularray}
\usepackage{booktabs}
\usepackage{wrapfig}
\usepackage{booktabs}
\usepackage{makecell}
\usepackage{multirow}
\usepackage{booktabs}

\usepackage{color,soul}
\definecolor{lightgray}{gray}{.9}

\usepackage{etoolbox}
\usepackage[normalem]{ulem}
\usepackage{subfigure}
\DeclareMathAlphabet{\mathcal}{OMS}{cmsy}{m}{n}
\DeclareMathOperator*{\sign}{sign}

\makeatletter
\patchcmd{\@makecaption}
{\scshape}
{}
{}
{}
\makeatother

\usepackage[cmintegrals]{newtxmath}

\makeatletter
\newcommand\fs@betterruled{%
  \def\@fs@cfont{\bfseries}\let\@fs@capt\floatc@ruled
  \def\@fs@pre{\vspace*{5pt}\hrule height.8pt depth0pt \kern2pt}%
  \def\@fs@post{\kern2pt\hrule\relax}%
  \def\@fs@mid{\kern2pt\hrule\kern2pt}%
  \let\@fs@iftopcapt\iftrue}
\floatstyle{betterruled}
\restylefloat{algorithm}
\makeatother

\usepackage{fancyhdr}
\begin{document}

\title{Logit Calibration and Feature Contrast for Robust Federated Learning on Non-IID Data}

\author{Yu Qiao,~\IEEEmembership{Student Member, IEEE},
        Chaoning Zhang,~\IEEEmembership{Member, IEEE},
        Apurba Adhikary,~\IEEEmembership{Student Member, IEEE},  \\
        and Choong Seon Hong,~\IEEEmembership{Fellow, IEEE}
\IEEEcompsocitemizethanks{

\IEEEcompsocthanksitem Yu Qiao and Chaoning Zhang are with the Department of Artificial Intelligence, School of Computing, Kyung Hee University, Yongin-si 17104, Republic of Korea (email: qiaoyu@khu.ac.kr;chaoningzhang1990@gmail.com).
\IEEEcompsocthanksitem Apurba Adhikary and Choong Seon Hong are with the Department of Computer Science and Engineering, School of Computing,
Kyung Hee University, Yongin-si 17104, Republic of Korea (e-mail: apurba@khu.ac.kr; cshong@khu.ac.kr).}}

\maketitle

\begin{abstract}
Federated learning (FL) is a privacy-preserving distributed framework for collaborative model training on devices in edge networks. However, challenges arise due to vulnerability to adversarial examples (AEs) and the non-independent and identically distributed (non-IID) nature of data distribution among devices, hindering the deployment of adversarially robust and accurate learning models at the edge. While adversarial training (AT) is commonly acknowledged as an effective defense strategy against adversarial attacks in centralized training, we shed light on the adverse effects of directly applying AT in FL that can severely compromise accuracy, especially in non-IID challenges. Given this limitation, this paper proposes FatCC, which incorporates local logit \underline{C}alibration and global feature \underline{C}ontrast into the vanilla federated adversarial training (\underline{FAT}) process from both logit and feature perspectives. This approach can effectively enhance the federated system's robust accuracy (RA) and clean accuracy (CA). First, we propose logit calibration, where the logits are calibrated during local adversarial updates, thereby improving adversarial robustness. Second, FatCC introduces feature contrast, which involves a global alignment term that aligns each local representation with unbiased global features, thus further enhancing robustness and accuracy in federated adversarial environments. Extensive experiments across multiple datasets demonstrate that FatCC achieves comparable or superior performance gains in both CA and RA compared to other baselines.
\end{abstract}


\IEEEpeerreviewmaketitle

\section{Introduction}
\IEEEPARstart{M}{obile} edge computing (MEC) is propelling the shift from traditional cloud computing to the edge network in next-generation computing networks~\cite{xiong2018mobile}. By deploying computing and storage capabilities at the edge, MEC establishes a node-edge-cloud architecture to support various applications on resource-constrained edge devices~\cite{wu2022node}. However, the proliferation of sensors, smartphones, and Internet of Things (IoT) devices is leading to a substantial increase in the data generated~\cite{ahmed2017role}. Meanwhile, the conventional practice of offloading data to edge servers for artificial intelligence (AI)~\cite{adhikary2023artificial_noms} model training raises concerns about data privacy~\cite{wang2023security}. Recently, an innovative distributed training strategy, known as federated learning (FL)~\cite{mcmahan2017communication}, has been proposed to address this concern. In FL, clients at the edge collaborate with a central server to train a shared global model while the raw data remains stored locally on the devices. Within this typical FL framework, several rounds of communication are performed until the global model converges, which includes global model distribution, local model training, model parameter transmission, and redistribution after global model averaging~\cite{qiao2023mp}. Throughout the iteration, the edge server can train a shared global model that each device can adopt without accessing its sensitive data. This distributed training method not only protects data privacy but also facilitates collaborative training to obtain a well-generalized global model~\cite{mcmahan2017communication}. Consequently, it is expected to be a highly promising technology in the field of edge computing.

Nonetheless, recent research has revealed parallel vulnerabilities observed in neural networks, echoing previous findings that, similar to models trained centrally, models undergoing an FL process are also susceptible to adversarial examples (AEs)~\cite{zizzo2020fat,hong2021federated,lyu2022privacy}. In particular, the attacker can cause highly inaccurate predictions (i.e., almost zero accuracy) by adding well-crafted and imperceptible perturbations to test samples during global model inference~\cite{zizzo2020fat,qiao2024noms_fedalc}. This raises significant security and reliability concerns when implementing FL in real-world scenarios. For example, in the domain of autonomous driving, a non-robust global model may inaccurately interpret traffic signs, consequently posing a risk of accidents~\cite{rossolini2023real}. Additionally, within the financial domain, vulnerable global models can result in misguided risk assessments or trading decisions, potentially leading to financial losses~\cite{yang2019federated}. Given these security and reliability concerns, it is imperative to design a robust FL model capable of defending against various adversarial attacks.

In centralized model training, adversarial training (AT) has emerged as one of the prevalent strategies to defend against adversarial attacks~\cite{goodfellow2014explaining}. Remarkably, the method based on projected gradient descent (PGD) attacks proposed in~\cite{madry2017towards} has emerged as one of the mainstream approaches for AT. This method is characterized by formulating a min-max optimization problem: within the inner loop, the objective is to craft the perturbation that maximizes the loss function, while within the outer loop, the model is then trained to minimize the loss on the AEs generated by this perturbation. In other words, the model is trained by incorporating AEs into its training process, thereby enhancing its resilience in the face of adversarial attacks. Fortunately, recent research~\cite{zizzo2020fat,li2023federated,qiao2024noms_fedalc} indicates that AT not only can enhance the robustness of models in centralized training environments but also exhibit potential in federated environments. Specifically, to address the security and reliability vulnerabilities that may exist in FL deployment, researchers initially introduce AT strategies into FL and terme it federated adversarial training (FAT)~\cite{qiao2024noms_fedalc,zizzo2020fat,hong2023federated,shah2021adversarial}. The difference between robust FAT and non-robust federated training (a.k.a vanilla FL in this paper) lies in the local update process, wherein FAT enhances global adversarial robustness by integrating PGD-based AT into local model training. However, although these methods can improve the robust accuracy (RA) of the global model, they tend to have relatively lower clean accuracy (CA) when making inferences on unperturbed samples using adversarially trained models~\cite{zizzo2020fat,qiao2024noms_fedalc}. Moreover, given the non-independent and non-identically distributed (non-IID) nature, which is widely prevalent in vanilla FL, this non-IID challenge still presents a significant challenge to the FAT framework. This difficulty makes it challenging to train a global model efficiently capable of simultaneously achieving high accuracy and robustness. We will clarify the differences arising from the non-IID challenge in vanilla FL and FAT environments in Section~\ref{sec:method}.

Motivated by the limitations mentioned above in previous studies, this paper focuses on enhancing both CA and RA when faced with adversarial attacks and non-IID challenges within an FL framework. First, to enhance the adversarial robustness of the federated system, we follow the common practice of integrating AT into local model updates. However, due to non-IID challenges, the direct adoption of the AT strategy in FL may still face the issue of low RA~\cite{luo2021ensemble}. Inspired by the long-tail learning method~\cite{menon2020long}, we propose a class frequency-based logit (i.e., the output of the last layer and the input of softmax) calibration strategy for the local AT process, aiming to mitigate local biases in achieving adversarial robustness. This calibration strategy, different from those in~\cite{chen2022calfat,zhang2022federated}, employs a modulating factor for enhanced flexibility without necessitating prior knowledge of class distributions. It can dynamically balance the sample distribution by assigning higher weights to the minority class and lower weights to the majority class within each mini-batch. Second, since each client optimizes towards a different local minimum, relying solely on its guidance signals makes global model optimization inconsistent and unreliable~\cite{huang2023federated}. Therefore, we construct unbiased global signals and further introduce the global alignment term that makes each local representation consistent with the global signals belonging to the same semantics while staying away from those with different semantics. We conjecture that combining these two components makes FatCC a competitive method for robust FL with non-IID data. Notably, the feature we utilize for communication is privacy-friendly, being only one dimension and undergoing two averaging operations~\cite{qiao2023mp,tan2022fedproto}. The main contributions of this paper are as follows:
\begin{itemize} 
\item[$\bullet$] We clarify that directly adopting the AT strategy to improve adversarial robustness in vanilla FL frameworks may have limited improvements in both CA and RA, especially in non-IID challenges.
\item[$\bullet$] We propose an effective algorithm termed FatCC, which involves calibrating the local AT process by adjusting logits and introducing a global alignment term based on feature contrast, to enhance both RA and CA within an adversarial federated framework.
\item[$\bullet$] Experimental results on three popular benchmark datasets, MNIST~\cite{lecun1998gradient}, Fashion-MNIST~\cite{xiao2017fashion}, and CIFAR-10~\cite{krizhevsky2009learning}, demonstrate that our approach is more competitive in terms of both CA and RA compared to several baselines.
\end{itemize}

The remainder of this paper is organized as follows. Related work is presented in Section \ref{sec:related_work}. The notation and preliminaries are provided in Section~\ref{sec:preliminary}. The methodology is presented in Section~\ref{sec:method}. Experimental results are provided in Section~\ref{sec:experiments}. Finally, conclusions are drawn in Section~\ref{sec:conclusion}.

\section{Related Work}
\label{sec:related_work}
In this section, we first review existing efforts to address challenges in FL in Section~\ref{subsec:fl}. Next, we discuss some popular contrastive learning techniques in Section~\ref{subsec:CL}. Finally, we provide works exploring adversarial attacks and defense for neural networks in Section~\ref{subsec:adv_attack}.
\subsection{Federated Learning}
\label{subsec:fl}
The concept of FL is initially introduced by McMahan~\cite{mcmahan2017communication}. Its representative algorithm, FedAvg~\cite{mcmahan2017communication}, embodies a classic distributed machine learning approach where multiple decentralized devices collaborate to protect local data privacy in model training. However, system heterogeneity and statistical heterogeneity typically exist among distributed devices~\cite{wen2023survey,ye2023heterogeneous}. Consequently, addressing system and statistical heterogeneity (a.k.a non-IID data) challenges has been a significant focus on the FL community since then. When dealing with the first challenge (i.e., system heterogeneity), efforts are focused on balancing computing power and storage resources variations between different devices. For example, FedAT~\cite{chai2021fedat} proposes an asynchronous layer where edge devices are grouped based on their system-specific capabilities. Sageflow~\cite{park2021handling} introduces a robust FL framework to tackle straggler issues. Additionally, CDFed~\cite{qiao2023cdfed} suggests a logical layer for grouping distributed devices according to their capabilities, thus minimizing the impact of hardware differences. To overcome the non-IID data challenge, various existing works~\cite{le2023fedmekt,tan2022fedproto,qiao2023mp,zhao2023ensemble,qiao2023knowledge} are dedicated to solving it from different perspectives. Approaches such as MP-FedCL~\cite{qiao2023mp} and FedProto~\cite{tan2022fedproto} propose maximizing prototype-level agreement between local and global models, thereby mitigating bias in local models towards their specific data distributions. Meanwhile, several other works~\cite{wang2020tackling,t2020personalized,karimireddy2020scaffold,yao2019federated,li2021model,li2020federated} delve into the incorporation of a regularization term into local models to address model bias. This strategy ensures that the update direction of each local model remains consistent with the global model. For instance, FedProx~\cite{li2020federated} proposes leveraging global model parameters as a reference to guide local model parameters closer to the global model during federated training. MOON~\cite{li2021model} adopts a similar approach but employs contrastive learning, further enhancing performance. SCAFFOLD~\cite{karimireddy2020scaffold} introduces a pair of control variables designed to capture updated directional information from both global and local models, effectively addressing gradient inconsistencies. Additionally, PFedMe~\cite{t2020personalized} utilizes the Moreau envelope function to decouple personalized and global model optimization models. This allows pFedMe to update the global model like FedAvg while optimizing each device's personalized model based on its local data. However, none of these works consider the adversarial robustness of FL models under adversarial attacks, which is more critical when deployed securely in the real world. In this paper, we focus on adversarial attacks and non-IID challenges, proposing a local logit calibration strategy and a global feature contrast term to learn a robust and accurate global model in the federated adversarial learning process.

\begin{table}[!t]
\caption{Summary of Notations.}
\label{tab: notations}
\centering
\begin{tabular}{|c||l|}
\hline
Notation & Description\\
\hline
FGSM & Fast gradient sign method attack \\
PGD & Projected gradient descent attack\\
BIM & Basic iterative method attack\\
AA & AutoAttack\\
CA & Clean accuracy \\
RA & Robust accuracy \\
$N$ & Number of distributed clients\\
$\mathcal{D}_i$ & Privacy-sensitive dataset for each client\\
$D_i$ & Size of $\mathcal{D}_i$ owned by each client\\
$\omega$ & Shared model parameters \\
$\boldsymbol {x}_i$ & Model input for each client\\
$y_i$ & Ground truth label for each client\\
$f_i(\omega; \boldsymbol {x}_i)$ & Local model for each client \\
$z_i$ & Logit output for each local model \\
$\mathbbm{1}(\cdot)$ & Indicator function\\
$\eta$ & Learning rate\\
$\nabla \mathcal{L}_i(\cdot)$ & Gradient of model parameters for each client \\
$\nabla \mathcal{L}(\omega_{t})$ & Gradient of the shared global model \\
$\delta$ & Perturbation for finding AEs \\
$\widetilde{\boldsymbol{x}}$ & AEs \\
$\mathcal{L}^{AT}_i(\omega)$ & Local AT for each client \\
$n_j$ & Number of samples for $j$-th class in a batch \\
$B$ & Size of each batch \\
$p_{i,j}$ & Probability of $j$-th class for client $i$ in a batch \\
$\alpha, \beta$ & Tunable parameters for modulating factor \\
$w_i$ & Weight for logit calibration \\
$f_{adv}^e(\boldsymbol {x}_{i,j})$ & Feature extractor module \\
$\mathcal{H}_{i,j}$ & Output of feature extractor module \\
$C_{i,j}$ & Size of class $j$ for each client \\
$\mathcal{Z}_{i,j}$ & Local feature of client $i$ belonging to $j$-th class \\
$\mathcal{G}$ & Global feature set \\
$\mathcal{P}_i$ & Positive samples set in global feature set \\
$\mathcal{K}_i$ & Negative samples set in global feature set \\
$\tau$ & Temperature for contrastive learning \\
$\mathcal{L'}^{{AT}}_i$ & Overall local objective for each client \\
$\mathcal{L'}_{{AT}}$ & Global objective for FatCC framework \\
\hline
\end{tabular}
\end{table}

\begin{figure*}[t]
\centering
\includegraphics[width=\textwidth]{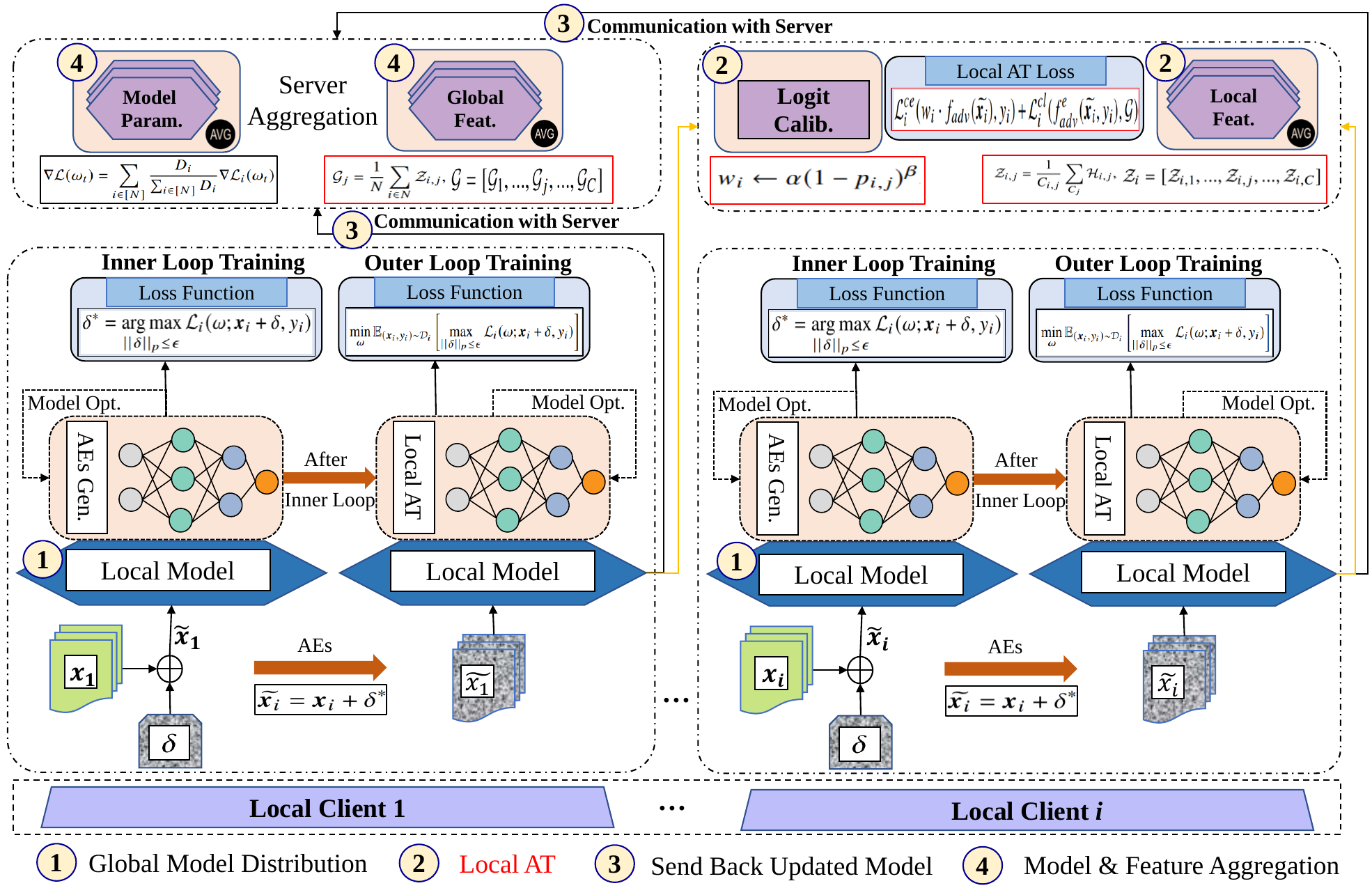}
\caption{Overview of the proposed FatCC training process. The main difference from the standard FL is mainly in the local training stage (i.e., step 2). During the local AT stage, we introduce a local logit calibration strategy to enhance the adversarial robustness of the local model (Sec.~\ref{subsec:local_re_weight}). Besides, we propose a global alignment term based on feature contrast to provide a consistency signal for further accuracy improvement (Sec.~\ref{subsec:global_regular}). }
\label{fig:system_model}
\vspace{-5px}
\end{figure*}

\subsection{Contrastive Learning}
\label{subsec:CL}
Contrastive learning~\cite{hadsell2006dimensionality} is a paradigm in self-supervised learnings~\cite{liu2021self} that has received widespread attention due to its ability to learn powerful representations without labeled data guidance. The purpose of this training method is to distinguish between positive pairs (similar samples) and negative pairs (dissimilar samples) within a dataset. The core practice of its design is to encourage the model to map similar samples close to each other while pushing dissimilar ones apart in the learned representations. To generate quality representations, contrastive learning methods rely on the number of negative samples. InstDis~\cite{wu2018unsupervised} is a seminal work that conducts contrastive learning between each instance and incorporates a memory bank strategy for storing negative sample features. However, maintaining the memory bank is memory-intensive and may limit learning effectiveness since only a subset of features in the memory bank can be updated after each mini-batch, while the model undergoes continuous updates. Following this, MoCo~\cite{he2020momentum} overcomes these limitations by introducing a dynamic dictionary with a queue and moving average encoders, which improves the effectiveness of contrastive learning by building a large but consistent dictionary in real-time. Moreover, SimCLR~\cite{chen2020simple} simplifies existing contrastive learning frameworks that rely on specialized architectures or memory banks, emphasizing the importance of constructing positive and negative pairs through a strategic composition of image augmentation methods. Subsequently, contrastive learning techniques have proven effective in various domains, including graph~\cite{you2020graph}, video~\cite{kuang2021video}, and audio~\cite{spijkervet2021contrastive, saeed2021contrastive}. Additionally, in the field of few-shot learning (FSL), some works~\cite{tan2022federated,li2021model,mu2023fedproc} also prove the effectiveness of contrastive learning in dealing with non-IID challenges. Unlike previous works, we construct contrastive learning within the FAT framework, aiming to improve the robustness and accuracy of the global model under adversarial attacks and non-IID challenges.

\subsection{Adversarial Attack and Defense}
\label{subsec:adv_attack}
Deep neural networks (DNNs), such as convolutional neural networks (CNN)~\cite{goodfellow2014explaining} and vision transformers (ViT)~\cite{dosovitskiy2020image}, are vulnerable to AEs~\cite{szegedy2013intriguing}, which are usually crafted by adding imperceptible perturbations to input images. The phenomenon of neural networks being sensitive to such small perturbations is identified in the seminal work~\cite{szegedy2013intriguing}, laying the foundation for research on adversarial attacks. Adversarial attacks can be categorized into two types: white-box attacks~\cite{qiao2023robustness} and black-box attacks~\cite{wang2023adversarial}. In white-box attacks, the attacker has access to the model structure and parameters, while in black-box attacks, such information is unavailable to the attacker. Fast gradient sign method (FGSM)~\cite{goodfellow2014explaining} is the first method to generate AEs, while PGD~\cite{madry2017towards} and basic iterative methods (BIM)~\cite{kurakin2018adversarial} can be recognized as an iterative version of FGSM. Subsequently, many variants have been designed for crafting AEs for stronger attacks, such as Square~\cite{andriushchenko2020square}, Carlini and Wagner (C\&W)~\cite{carlini2017towards}, and AA~\cite{croce2020reliable} attacks. In addition to image-dependent perturbations, researchers have also found the existence of image-independent universal attack perturbations (UAPs)~\cite{moosavi2017universal}, which can cause the neural network to misclassify all images. To defend against adversarial attacks, various methods, including input or model modifications and the incorporation of external modules~\cite{akhtar2021advances}, have been devised, but AT stands out as the most recognized and effective defense strategy~\cite{zhang2019limitations}. Recently, given the issue of secure deployment, several works~\cite{zizzo2020fat,qiao2024noms_fedalc,hong2023federated,chen2022calfat} have already applied AT in FL to obtain a robust global model. For example,~\cite{zizzo2020fat,hong2023federated} focus on RA improvement by conducting AT on a proportion of clients. ~\cite{chen2022calfat} proposes to reweight each client's logit output based on the prior probability of the class, but considering the privacy-preserving nature of the FL environment, this approach may violate its inherent limitations. Unlike the existing works, in this paper, we propose to improve the CA and RA by performing local logit calibration and global feature contrast without violating the constraints of privacy protection in FL environments.

\section{Notation and Preliminaries}
\label{sec:preliminary}
In this section, we first introduce the basic setup of standard FL in Section~\ref{subsec:notation_fl}, followed by a discussion on basic AEs generation in Section~\ref{subsec:notation_adv_attack}. Subsequently, we present adversarial training techniques applied in FL scenarios in Section~\ref{subsec:adv_training}.

\subsection{Federated Learning}
\label{subsec:notation_fl}
The FL framework aims to achieve a well-trained shared global model through collaboration between distributed clients and an edge server, ensuring local client data privacy protection. The training process can be summarized as follows:

Consider a federated environment involving $N$ distributed devices and an edge server. Each device, denoted as $i$, possesses its private and sensitive dataset $\mathcal{D}_i$ consisting of image-label pairs represented as $\boldsymbol{x}_i$ and $y_i$, respectively. 
The size of the dataset owned by each device is denoted as $D_i$. The objective is to train a shared model for each client through cooperation between clients and the edge server. We denote the model output (logits) for each client as $z_i = f_i(\omega; \boldsymbol {x}_i)$. Then the cross-entropy loss for each client $i$ with one-hot encoded labels can be defined as follows~\cite{qiao2023mp}:
\begin{equation} \label{one_hot}
   {f_i} (\omega) = -\sum_{j=1}^{C}\mathbbm{1}_{y = j}\log \frac{exp(z_{i,j})}{exp(\sum_{j=1}^{C}{z_{i,j}})},
\end{equation}
where $\mathbbm{1}(\cdot)$ denotes the indicator function, and $\omega$ represents the shared model parameters of the global model. The discrete label set $[C]$ encompasses $C$ classes, where $C$ denotes the number of classes. The local loss for each client $\mathcal{L}_i$ can be given as:
\begin{equation} \label{local_loss}
   {\mathcal{L}_i} (\omega) = \frac {1}{D_i} \sum_{i \in \mathcal{D}_i} f_i(\omega).
\end{equation}

At the global round $t+1$, each local client joins the FL training and performs local stochastic gradient descent (SGD) to update its local weights. Formally,
\begin{equation} \label{local_sgd}
 \omega_{t+1}^i  =  \omega_{t} - \eta \nabla \mathcal{L}_i(\omega_{t}),
\end{equation}
where $\eta$ represents the learning rate, $\nabla \mathcal{L}_i(\cdot)$ is the local gradient of client $i$, and $\omega_{t}$ denotes the updated parameters of the global model from the previous round. 

Then, the global objective is to calculate the local loss across distributed clients as follows:
\begin{equation} \label{global_objective}
    \mathcal{L} (\omega) = \sum_{i \in [N]} \frac {D_i}{\sum_{i \in [N]}D_i} {\mathcal{L}_i} (\omega),
\end{equation}
where $[N]$ denotes the set of distributed clients with $[N] = \{1,...,N\}$, and the global gradient is calculated as follows:
\begin{equation} \label{global_weight}
\nabla \mathcal{L}(\omega_{t}) = \sum_{i \in [N]}\frac {D_i}{\sum_{i \in [N]}D_i} \nabla \mathcal{L}_i(\omega_{t}).
\end{equation}

Then, the global weights are updated at the global round $t+1$, as follows: 
\begin{equation} \label{global_sgd}
 \omega_{t+1}  =  \omega_{t} - \eta \nabla \mathcal{L}(\omega_{t}).
\end{equation}

Overall, the objective is to minimize the global loss during the FL process, as follows:
\begin{equation} \label{global_loss}
   \min_{\omega} {\mathcal{L}(\omega)}.
\end{equation}

\subsection{AEs Generation}
\label{subsec:notation_adv_attack}
Adversarial attacks aim to find AEs that can fool a trained model. These examples are generated by deliberately adding invisible perturbations to the input data with the goal of causing the model to make incorrect predictions. Considering a dataset from a certain client, without loss of generality, we denote its image classifier as $g(\omega; \boldsymbol{x}_i): \mathbb{R}^{h \times w \times c}\rightarrow [C]$. This classifier maps the input image $\boldsymbol{x}_i$ to a discrete label set $[C]$ with $C$ classes, where $h$, $w$, and $c$ denote the image's height, width, and channel, respectively. The adversary aims to find a perturbation $\delta \in \mathbb{R}^{h \times w \times c}$ that maximizes the loss function $\mathcal{L}_i(\omega; \boldsymbol{x}_i)$ for each client, resulting in $g(\boldsymbol{x}_i + \delta) \neq g(\boldsymbol{x}_i)$. Therefore, the optimal perturbation $\delta^{\ast}$ can be optimized as follows~\cite{bai2021recent}:
\begin{equation} \label{delta_optimi}
   \delta^{\ast} = \mathop {\arg\max}_{||\delta||_p \leq \epsilon} \mathcal{L}_i(\omega; \boldsymbol{x}_i + \delta, y_i),
\end{equation}
where $\epsilon$ denotes an upper bound on $\ell_p$-norm so that the perturbation $\delta$ is imperceptible (or quasi-imperceptible) to human eyes, and $p$ can be 0, 1, 2, or $\infty$ based on different algorithms. Then, the AEs for each client, $\widetilde{\boldsymbol{x}}$, can be expressed as follows:
\begin{equation} \label{AE_gen}
   \widetilde{\boldsymbol{x}_i} = \boldsymbol{x}_i + \delta^{\ast}.
\end{equation}

\subsection{Adversarial Training}
\label{subsec:adv_training}
An effective and widely recognized method to defend against adversarial attacks is AT~\cite{athalye2018robustness}. Its purpose is to build an adversarially robust model that can generalize well to any small perturbations added to the input data. In particular, the method formalizes the problem as a min-max problem by minimizing the prediction error against an adversary that interferes with the input and maximizes adversarial loss. Inspired by the success of AT in centralized training, the FL community has adopted a similar approach, performing AT in the local update process~\cite{qiao2024noms_fedalc}, with the objective of enhancing the robustness of the global model. Consequently, the local AT for each client $\mathcal{L}^{AT}_i$ can be formulated below:
\begin{equation} \label{AT_federated}
   \mathop {\min}_{\omega} \mathbb E_{(\boldsymbol{x}_i, y_i) \sim \mathcal{D}_i} \left[\mathop {\max}_{||\delta||_p \leq \epsilon} \mathcal{L}_i(\omega; \boldsymbol{x}_i + \delta, y_i)\right],
\end{equation}
where the inner maximization problem involves finding the most challenging samples for each local client, while the outer minimization problem aims to optimize the model's robustness against the found AEs.

The most common solution to the inner problem is a multi-step gradient-based attack, typically generated through the PGD attack as follows:
\begin{equation}
\label{eq:pgd}
\boldsymbol x_i^{t+1}=\Pi_{\boldsymbol{x_i} + \delta}\left(\boldsymbol x_i^t+\alpha\sign(\nabla_{\boldsymbol x^t}  \mathcal{L}_i(\omega; \boldsymbol{x}_i^t, y_i) \right),
\end{equation}
where $\alpha$ represents the step size, $\boldsymbol x_i^t$ is the AE generated at $t$-th step, $\Pi_{\boldsymbol x_i + \delta}$ denotes the projection function that projects the AE onto the $\epsilon$-ball centered at $\boldsymbol x_i^0$, and $\sign(\cdot$) indicates the sign function. Note that in order to ensure that the perturbation $\delta$ is imperceptible (or quasi-imperceptible) to the human eye, it is usually constrained by an upper bound $\epsilon$ on the $\ell_\infty$-norm, i.e., $||\delta||_\infty \leq \epsilon$.

\begin{figure}[t]
\centering
\includegraphics[width=0.40\textwidth]{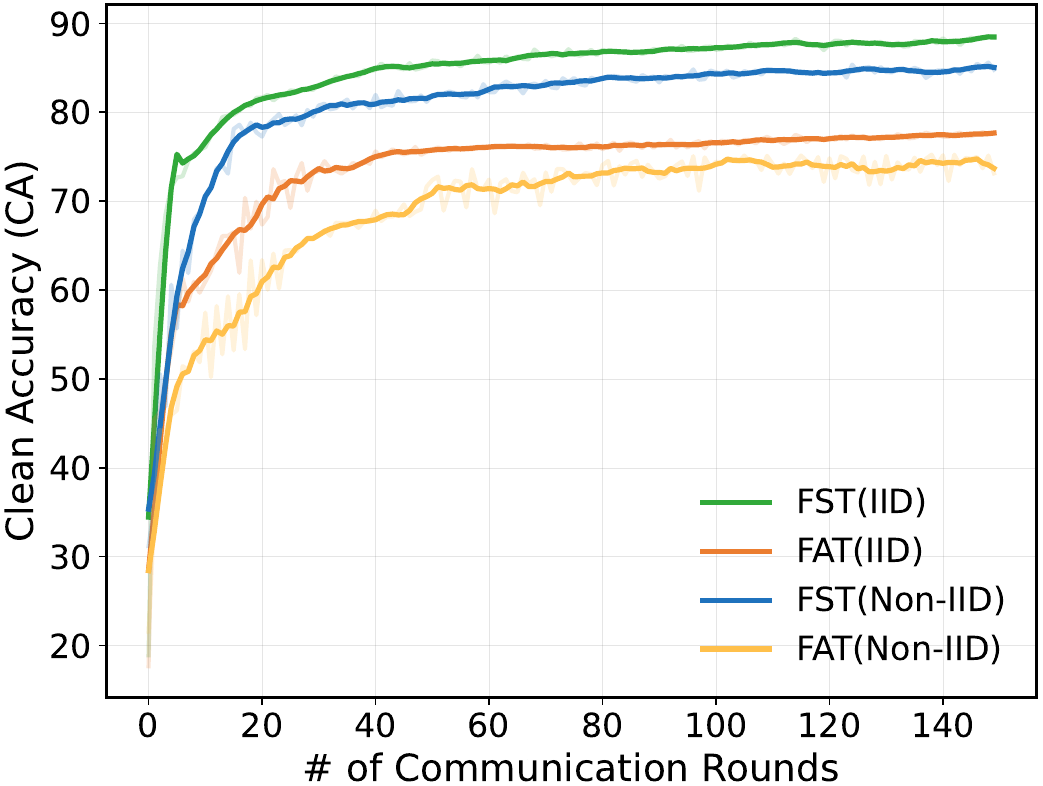}
\caption{CA (\%) comparison between FST and FAT strategies under both IID and non-IID scenarios with Fashion-MNIST dataset.}
\label{fig:tease_example}
\end{figure}

\begin{table}[t]
\centering
\caption{Comparison of CA (\%) and RA (\%) of MNIST based on non-IID setting under AA attack, the perturbation level is 0.3~\cite{goodfellow2014explaining}.}
\label{tab:tease_example}
\vspace{-5px}
\resizebox{0.35\textwidth}{!}{
\begin{tabular}{ccccc}
\toprule
Algo & FST  & FAT  & FatCC (ours)\\
\midrule
CA & 91.38 & 72.96 & 96.74 \\
RA (AA) & 0 & 4.44 & 23.38 \\
\bottomrule
\end{tabular}}
\end{table}

After finishing each local training during every global iteration, each client uploads its adversarially trained model parameters to the server for aggregation, a process known as FAT. Then, the overall objective in Eq.~\ref{global_loss} can be reformulated as below:
\begin{equation} \label{eq:global_at}
    \mathop {\min}_{\omega} {\mathcal{L}_{AT}} (\omega) = \sum_{i \in [N]} \frac {D_i}{\sum_{i \in [N]}D_i} {\mathcal{L}^{AT}_i} (\omega; \boldsymbol{x}_i, y_i),
\end{equation}
where $\mathcal{L}_{AT}(\omega)$ and $\mathcal{L}^{AT}_i(\omega)$ represent the local and global adversarial training loss, respectively. This formulation involves performing AT on each client in its local updates to enhance its robustness, thus contributing to the overall enhancement of the global model's robustness after federated training. In this paper, we follow this strategy but focus on optimizing the local AT process through local calibration and global feature contrast strategies to improve both CA and RA. We provide a summary of notations used in this paper in Table~\ref{tab: notations}.

\section{Methodology}
\label{sec:method}
In this section, we first clarify the adverse effect posed by non-IID challenges to FAT in Section~\ref{subsec:non_IID_FAT}, where FAT is a framework directly employing AT within FL. Next, we propose the FatCC framework in Section~\ref{subsec:proposed_framework}. Subsequently, we propose two strategies for local model AT training in Section~\ref{subsec:local_re_weight} and Section~\ref{subsec:global_regular}. Finally, we propose the overall objective in Section~\ref{subsec:overall_obj}.

\subsection{Non-IID Challenges in FAT framework}
\label{subsec:non_IID_FAT}
We clarify the non-IID challenge by presenting the CA and RA results, comparing the vanilla FL and FAT on the Fashion-MNIST dataset~\cite{xiao2017fashion} in both IID and non-IID scenarios, as depicted in Figure~\ref{fig:tease_example}. For a clear distinction between vanilla FL and the FAT framework, we employ the term federated standard training (FST) to denote the former. From these results, several observations can be made. First, in the IID scenario, FAT shows significant performance degradation compared to FST, indicating that AT may have a negative impact on model performance. Second, in the non-IID scenario, the performance of both FST and FAT decreases, but the decrease of FAT is more significant. Third, from an overall trend, FAT shows lower CA regardless of whether the data is IID, highlighting the performance challenges introduced by naively using AT in FL, especially in the non-IID scenario. Table~\ref{tab:tease_example} further highlights the challenges through a quantitative comparison of AutoAttack (AA)~\cite{croce2020reliable} attack on MNIST~\cite{lecun1998gradient}. The results confirm the challenges posed by adversarial attacks, as evidenced by the 0 RA after the AA attack. Moreover, the reduction in CA after the AT process (i.e., FAT) compared to FST, and the competitive performance of FatCC in RA compared to FAT, highlight the limited effect of direct adoption of AT in FL.

\subsection{Proposed FatCC Framework}
\label{subsec:proposed_framework}
The concept of FAT is initially introduced by~\cite{zizzo2020fat} as a solution to resist the vulnerability of FL on AEs. We follow this framework and propose the FatCC framework, whose primary training process is shown in Figure~\ref{fig:system_model}. For simplicity, only one global iteration is marked in this figure. Similar to the standard FL process, FatCC involves four main steps. First, the server sends the initialized global model to the distributed clients (step 1). Second, each local client updates the received model parameters in an AT manner based on its local dataset (step 2). Third, all participating devices return their updated model parameters to the server for aggregation (step 3). Finally, all received model parameters are aggregated at the server (step 4), repeating these steps until convergence. We focus on the second stage, where each local model is trained using the AT strategy. In detail, during the local training phase of each device, an imperceptible perturbation $\delta$ is added to the input data $\boldsymbol {x}_i$, thereby generating AE $\widetilde{\boldsymbol {x}}_i $. Subsequently, each device needs to optimize its local model parameters to resist this adversarial perturbation while maintaining a good CA. The details of our proposed local AT process are illustrated in the following sections.

\subsection{Local Calibration with Logit Adjustment}
\label{subsec:local_re_weight}
Within the AT paradigm, the neural network architecture $f_{adv}(\widetilde{\boldsymbol{x}})$ can be divided into two main components: the feature extraction layer $f_{adv}^e(\widetilde{\boldsymbol{x}})$ and the classification layer $g_{adv}(\widetilde{\boldsymbol{x}})$. The former serves the role of mapping the input space to an embedded space, and the responsibility of the classification layer lies in mapping the embedded space to a logit space. By comparing the predicted logits with the ground truth labels, the model parameters are iteratively updated to minimize the loss, thereby enhancing the model's accuracy. However, as previously mentioned, the data distribution among distributed clients in the FL framework is typically non-IID. The number of instances for each class varies among different clients. Directly updating based on clients' biased local data distribution may introduce biases towards the majority classes, especially within the FAT environment. We note that this paper focuses on the representative label non-IID setting~\cite{mu2023fedproc}, where the label distribution varies, while the feature distribution is similar for all clients. As revealed by~\cite{menon2020long}, logit adjustment based on class occurrence probabilities proves advantageous in alleviating label distribution bias. Motivated by this, it can be expected that by calibrating the logit before softmax cross-entropy based on each class's probability of occurrence, we can effectively alleviate the label distribution bias for each local model, at least to a certain extent, so that it does not be biased towards its majority classes.

Specifically, it is essential to assign greater weight to the logits of the minority classes and a smaller weight to the logits of the frequent classes, thereby better balancing the uneven label distribution. For an arbitrary client, let $n_j$ represent the number of samples of $j$-th class within a mini-batch sample, and the size of each batch is represented by $B$. Then, the probability of class occurrence~\cite{qiao2024noms_fedalc} can be defined as follows:
\begin{equation} \label{eq:frequency}
p_{i,j} = \frac{n_j}{B}, \quad i \in N,  \quad j \in [C],
\end{equation}
where $p_{i,j}$ denotes the probability of the $j$-th class for client $i$ within a batch. 

During training, the model tends to favor classes with higher occurrence probabilities. Therefore, intuitively, we should assign smaller weights to these high-probability classes, and vice versa. This approach enhances the balance in the learning process among different classes. More formally, we propose to add a weighted modulating factor~\cite{lin2017focal} as the weight of the logit value, which can be formulated as:
\begin{equation} \label{eq:re_weight}
w_i \gets \alpha (1 - p_{i,j})^\beta, \quad i \in N,  \quad j \in [C],
\end{equation}
where $w_i$ is the weight used for logit calibration of each client, and $\alpha > 0$ and $\beta \geq 0 $ are tunable parameters.

We observe three properties of the modulating factor. First, when a class has more samples (i.e., $p_{i,j}$ is close to 1), the factor approaches 0, leading to a down-weighting of the majority class. Conversely, when the class has fewer samples, the modulating factor increases, resulting in an up-weighting of the minority class. Second, $\alpha$ is used to scale the modulating factor. By adjusting the value of $\alpha$, we can control the degree of effect of the factor on the weight. Generally, a larger $\alpha$ may lead the modulating factor to have a more significant impact on the weight. Third, the parameter $\beta$ smoothly adjusts the rate at which majority classes are down-weighed and minority classes are up-weighted. When $\beta = 0$, the weight $w_i$ for each class is the same and is 1, and as $\beta$ is increased, the effect of the modulating factor is correspondingly increased. For example, when $\beta = 2$ and $\alpha = 1$, a class frequency probability of $p_{i,j} = 0.9$ would have a weight $81\times$ lower than that of a class frequency probability of $p_{i,j} = 0.1$. Further, with $\beta = 4$ and $\alpha = 1$, the weight assigned to a class frequency probability of $p_{i,j} = 0.9$ would be $6561\times$ lower than that of a class frequency probability of $p_{i,j} = 0.1$.

\subsection{Global Alignment with Feature Contrast}
\label{subsec:global_regular}
In general, the goal of FL is to acquire a shared global model based on locally biased data from different clients, demonstrating effective generalization capabilities when applied to unbiased test data. Nonetheless, a notable challenge arises from the intrinsic divergence of local models from the optimal global solution, presenting difficulties in the optimization process for the shared global model. Motivated by~\cite{li2021model,tan2022fedproto}, they reveal that the shared global model presents less bias than local models in a typical FL environment. We argue that averaged global features from multiple parties still should have less bias than local features in an adversarial federated environment. Before delving into the details of our global feature contrast loss design, we first give the method for the calculation of the local and global features in the FAT environment. Specifically, for the client $i$, its local feature of $j$-th class generated by feature extraction layer $\mathcal{H}_{i,j} = f_{adv}^e(\boldsymbol {x}_{i,j})$ during local AT process can be calculated as:
\begin{equation} \label{local_feature}
\begin{aligned}
 \mathcal{Z}_{i,j} &=\frac{1}{C_{i,j}} \sum_{C_j} \mathcal{H}_{i,j}, \quad i \in N, \quad j \in [C], \\
 \mathcal{Z}_i &= [\mathcal{Z}_{i,1},..., \mathcal{Z}_{i,j},..., \mathcal{Z}_{i,C}],
 \end{aligned}
\end{equation}
where $\mathcal{Z}_{i,j}$ represents the local feature of client $i$ corresponding to the $j$-th class, $\mathcal{Z}_i$ denotes the local feature set of client $i$, and $C_{i,j}$ is the size of class $j$ for client $i$. This formula aims to average the feature embedding belonging to the same class space for each client. Note that the local feature is only a one-dimensional vector; therefore, it has significantly fewer parameters than the original raw data.

Considering the communication-sensitive nature in the FL environment, a simple yet effective way to exploit local features is to derive global features through an averaging operation, which is similar to the generation of global models. Compared to local features, this global feature encapsulates knowledge from each client and has a relatively consistent optimization goal, which can be calculated as follows:

\begin{equation} \label{global_feature}
\begin{aligned}
\mathcal{G}_j &= \frac{1}{N} \sum_{i\in N} \mathcal{Z}_{i,j} , \quad i \in N, \quad j \in [C], \\
\mathcal{G} &= [\mathcal{G}_1,..., \mathcal{G}_j,..., \mathcal{G}_C],
\end{aligned}
\end{equation}
where $\mathcal{G}_j$ denotes the global feature of $j$-th class, and $\mathcal{G}$ denotes the set of all global features. This averaging process involves calculating the average of all clients from class $j$ with local features of that class. Note that the global feature is privacy-friendly because it is only a one-dimension vector and experiments twice averaging operation. 

During the local AT process, we expect that the local model remains unbiased not only towards the majority classes of its local dataset but also avoids deviations from the global optimum. Therefore, the utilization of global features serves as a guide for each local training, providing a consistent direction for iteration. Moreover, it is generally acknowledged that a highly generalized representation not only needs to maintain the ability to distinguish between different classes but also increase the semantic dispersion between them as much as possible. Building upon this understanding and drawing inspiration from the success of supervised contrastive learning~\cite{khosla2020supervised}, we introduce a method to federated adversarial environments that regularizes the direction for local AT updates by contrasting each client's local adversarial features with the global features, thereby further improving the robustness and accuracy. The objective is to pull the adversarial feature vector closer to positive samples with the same semantics as the global feature while simultaneously pushing them away from negative samples belonging to distinct classes. Our supervised contrastive loss within an adversarial federated environment for each client is defined as:
\begin{equation} \label{cl_loss}
    \mathcal L_i^{cl} = \frac{-1}{|\mathcal{P}_i|}\sum_{p \in \mathcal{P}_i}\log\frac{\psi(\mathcal{H}_i, \mathcal{G}_p, \tau)}{\psi(\mathcal{H}_i, \mathcal{G}_p, \tau) + \sum_{k \in \mathcal{K}_i}\psi(\mathcal{H}_i, \mathcal{G}_k, \tau)},
 \end{equation}
 where $\mathcal{P}_i$ and $\mathcal{K}_i$ denote the set of positive and negative samples in the global feature $\mathcal{G}$, respectively, $\tau$ is a temperature hyperparameter and $\psi$ is formulated as:
 \begin{equation}
    \psi(\mathcal{H}_i, \mathcal{G}_j, \tau) = \exp({\frac{\mathcal{H}_i, \mathcal{G}_j}{\|\mathcal{H}_i\| \times \| \mathcal{G}_j\|}} / \tau).
 \end{equation}

To better understand the behavior of contrastive learning in Eq.~\ref{cl_loss} within federated adversarial environments, we apply Taylor expansion and reformulate it as below:
\begin{equation}
    \begin{aligned} \label{decoupled_cl_loss}
        \mathcal L_i^{cl} 
        &= \frac{1}{|\mathcal{P}_i|}\sum_{p \in \mathcal{P}_i}\log(1 + \frac{\sum_{k \in \mathcal{K}_i}\psi(\mathcal{H}_i, \mathcal{G}_k, \tau)}{\psi(\mathcal{H}_i, \mathcal{G}_p, \tau)}), \\
        &\approx \frac{1}{|\mathcal{P}_i|}\sum_{p \in \mathcal{P}_i} \frac{\sum_{k \in \mathcal{K}_i}\psi(\mathcal{H}_i, \mathcal{G}_k, \tau)}{\psi(\mathcal{H}_i, \mathcal{G}_p, \tau)}, \\
        &= \frac{\mathcal L_i^{cl-}(\mathcal{H}_i, \mathcal{K}_i)}{\mathcal L_i^{cl+}(\mathcal{H}_i, \mathcal{P}_i)},
    \end{aligned}
\end{equation}
where we denote $\sum_{k \in \mathcal{K}_i}\psi(\mathcal{H}_i, \mathcal{G}_k, \tau)$ as $\mathcal{L}_i^{cl-}(\mathcal{H}_i, \mathcal{K}_i)$, representing the loss calculated on negative samples. Similarly, $\frac{1}{|\mathcal{P}i|}\sum_{p \in \mathcal{P}_i} \psi(\mathcal{H}_i, \mathcal{G}_p, \tau)$ is denoted as $\mathcal{L}_i^{cl+}(\mathcal{H}_i, \mathcal{P}_i)$, representing the loss calculated on positive samples.

In the reformulated loss Eq.~\ref{decoupled_cl_loss}, minimizing $\mathcal{L}_i^{cl}$ is equivalent to minimizing $\mathcal{L}_i^{cl-}(\mathcal{H}_i, \mathcal{K}_i)$ and maximizing $\mathcal{L}_i^{cl+}(\mathcal{H}_i, \mathcal{P}_i)$. Since contrastive loss typically involves cosine similarity, minimizing $\mathcal{L}_i^{cl-}(\mathcal{H}_i, \mathcal{K}_i)$ implies pushing the query sample $\mathcal{H}_i$ far away from the negative samples $\mathcal{K}_i$, while maximizing $\mathcal{L}_i^{cl+}(\mathcal{H}_i, \mathcal{P}_i)$ means pulling the query sample $\mathcal{H}_i$ closer to the positive samples $\mathcal{P}_i$. In other words, this objective aims to maintain semantic distance between different classes and to ensure robustness against samples from the same classes but originating from diverse sources. As a result, this adversarial contrastive learning approach offers both generalizable and discriminative properties, leading to satisfactory performance in adversarial federated environments.

\subsection{Overall Objective}
\label{subsec:overall_obj}
Our proposed adversarial FL framework mainly consists of two key components. First, we propose to calibrate the cross-entropy loss based on class frequency to improve the adversarial robustness of the model against AEs. Second, we propose a global consistency term based on feature contrast to improve the model's accuracy further. Then, the AT loss for each client in Eq.~\ref{AT_federated} can be rewritten as:
\begin{equation}\label{eq:modified_global_loss}
      \min \mathcal{L'}^{{AT}}_i = \underbrace{ \mathcal L_i^{ce}(w_i \cdot f_{adv}(\widetilde{\boldsymbol{x}_i}),y_i)}_{\text{logit calibration}} 
      + \underbrace{ \mathcal L_i^{cl}(f_{adv}^e(\widetilde{\boldsymbol{x}_i},y_i), \mathcal{G})}_{\text{feature contrast}},
\end{equation}
where $\mathcal{L'}^{{AT}}_i$ is the proposed local objective, $\mathcal L_i^{ce}$ represents the calibrated cross-entropy loss to improve adversarial robustness, and $\mathcal L_i^{cl}$ is the contrastive loss that further offers consistency for the local feature of each client with unbiased global features to improve accuracy. 

Finally, the overall objective of our proposed adversarial federated training framework is to optimize across distributed clients. Then, Eq.~\ref{eq:global_at} can be rewritten as follows:
\begin{equation} \label{eq:adv_global_loss}
    \mathop {\min}_{\omega} {\mathcal{L'}_{{AT}}} (\omega) = \sum_{i \in [N]} \frac {D_i}{\sum_{i \in [N]}D_i} {\mathcal{L'}^{{AT}}_i},
\end{equation}
where $\mathcal{L'}^{{AT}}$ is the proposed overall objective. A more detailed training process of FatCC is presented in Algorithm \ref{alg:FatCC}. The input to the algorithm is heterogeneous datasets and training parameters from different clients. When the federated system initialization is completed, the proposed FatCC training process is executed from line 2 to line 10. In each global iteration, all clients perform adversarial federated training in parallel from lines 3 to 6. For each client, the calculation of the modulating factor for the local logit calibration strategy is executed in line 15, followed by the completion of local feature calculation in line 17. Moreover, the global feature contrast loss calculation takes place in line 18. Finally, the computation of the local overall objective for each client is performed in line 20. After performing SGD for each local client in line 21, each client subsequently transmits its updated model parameters and computed local features in line 24 back to the server. The server then performs model parameter aggregation in line 9 and global feature aggregation in line 8, starting the next global iteration until the total global rounds $T$ are completed.

\begin{algorithm}[t] 
    \caption{FatCC} 
    \label{alg:FatCC} 
    \begin{algorithmic}[1]
        \REQUIRE ~~ \\
        Dataset $\mathcal{D}_i$ of each client, $\omega_i$, number of clients $N$.
        \STATE \textbf{Initialize $\omega^0$}.
        \FOR{ $t$ = 1, 2, ..., $T$} 
            \FOR{ $i$ = 0, 1,..., $N$ \textbf{in parallel}}
                \STATE Send global model $\omega^t$ to client \textit{i}
                \STATE {$\omega^t, \mathcal{Z}_{i} \gets \textbf{LocalUpdate}(\omega^t$)}
            \ENDFOR
            \STATE {\textcolor{gray}{\text{/* Global feature aggregation */}}}
            \STATE {$\mathcal{G}_j \gets \frac{1}{N} \sum_{i\in N} \mathcal{Z}_{i,j}$} via Eq.~\ref{global_feature}
            \STATE {$\nabla \mathcal{L}(\omega_{t}) \gets \sum_{i \in [N]}\frac {D_i}{\sum_{i \in [N]}D_i} \nabla \mathcal{L}_i(\omega_{t})$ by Eq.~\ref{global_weight}}
        \ENDFOR \\
        \textbf {LocalUpdate($\omega^t$, $\mathcal{G}$)}
        \FOR{ each local epoch }
        \FOR{ each batch ($\boldsymbol{x}_i$; $y_i$) of $\mathcal{D}_i$}
        \STATE {$\widetilde{\boldsymbol{x}_i} \gets \boldsymbol{x}_i + \delta^{\ast}$} by Eq.~\ref{AE_gen}
        \STATE {\textcolor{gray}{\text{/* Modulating factor calculation */}}}
        \STATE {$w_i \gets \alpha (1 - p_{i,j})^\beta$} via Eq.~\ref{eq:re_weight}
        \STATE{\textcolor{gray}{\text{/* Local feature calculation */}}}
        \STATE{$\mathcal{Z}_{i,j} \gets \frac{1}{C_{i,j}}\sum_{C_j} \mathcal{H}_{i,j}$} by Eq.~\ref{local_feature}\\
        \textcolor{gray}{\text{/* Feature contrast loss calculation */}}
        \STATE {$\mathcal L_i^{cl} \gets \frac{-1}{|\mathcal{P}_i|}\sum_{p \in \mathcal{P}_i}\log\frac{\psi(\mathcal{H}_i, \mathcal{G}_p, \tau)}{\psi(\mathcal{H}_i, \mathcal{G}_p, \tau) + \sum_{k \in \mathcal{K}_i}\psi(\mathcal{H}_i, \mathcal{G}_k, \tau)}$} by Eq.~\ref{cl_loss}
        \STATE \textcolor{gray}{\text{/* Local objective for each client */}} 
        \STATE $\mathcal{L'}^{{AT}}_i \gets \mathcal L_i^{ce}(w_i \cdot f_{adv}(\widetilde{\boldsymbol{x}_i}),y_i) + \mathcal L_i^{cl}(f_{adv}^e(\widetilde{\boldsymbol{x}_i},y_i), \mathcal{G})$ via Eq.~\ref{eq:modified_global_loss} \\
        \STATE {$ \omega_{t+1}  \gets  \omega_{t} - \eta \nabla \mathcal{L'}^{{AT}}_i$ via Eq.~\ref{local_sgd}}
    \ENDFOR
    \ENDFOR
        \RETURN $\omega_i^t$, $\mathcal{Z}_{i}$
    \end{algorithmic}
\end{algorithm}

\section{Experiments}
\label{sec:experiments}
In this section, we first introduce the experimental setup in Section~\ref{subsec:setup}. Next, the choice of hyperparameters will be discussed in Section~\ref{subsec:choose}. The accuracy and robustness comparison are shown in Section~\ref{subsec:accuracy_compar} and Section~\ref{subsec:robust_compar}, respectively. Finally, we conduct an ablation study in Section~\ref{subsec:ablation} to illustrate the effectiveness of each component in our proposed framework.

\subsection{Experimental Setup}
\label{subsec:setup}
\textbf{Datasets.} We conduct experiments for the proposed scheme on multiple popular benchmark datasets: MNIST~\cite{lecun1998gradient}, Fashion-MNIST~\cite{xiao2017fashion} and CIFAR-10~\cite{krizhevsky2009learning} to verify the potential advantages of FatCC for robust edge intelligence. Before delving into the detailed experimental results, we briefly introduce the datasets used. \textbf{MNIST} is a dataset for handwritten digit recognition, while \textbf{Fashion-MNIST} is a dataset consisting of 10 different categories of fashion items. Both MNIST and Fashion-MNIST have 10 distinct classes, with 60,000 training samples and 10,000 test samples for each. \textbf{CIFAR-10} poses a more challenging task, featuring 60,000 images across 10 categories, with 50,000 training images and 10,000 test images.

\textbf{Local model setup.} For MNIST and Fashion-MNIST model setup, we adopt a simple CNN model~\cite{qiao2024noms_fedalc,chen2022calfat} consisting of five layers, with the following structure: a 5x5 convolution layer, followed by a 2x2 max pooling layer, the other 5x5 convolution layer, followed by a 2x2 max pooling layer, and finally, followed by 3 fully connected layers. The ReLU activation function is applied after each convolutional layer and fully connected layer. Considering that CIFAR-10 is a more challenging task compared to MNIST and Fashion-MNIST, we opt for a deeper CNN model architecture, ResNet-18~\cite{he2016deep}. The feature vector dimension is 80 for both MNIST and Fashion-MNIST, while it is 512 for CIFAR-10. We note that for a fair comparison, all baselines follow the same model architecture.

\begin{table}[t]
\centering
\caption{Effect of hyper-parameters $\alpha$ and $\beta$ for MNIST, Fashion-MNIST, and CIFAR-10. The empirically chosen trade-off between CA and RA is in \textbf{bold}.}
\label{tab:alpha_beta_chosen_all}
\resizebox{\linewidth}{!}{
\begin{tabular}{lcccccccccc}
\toprule
\multirow{2}{*}{Dataset} & \multirow{2}{*}{$\beta$} & \multicolumn{2}{c}{$\alpha=1$} & \multicolumn{2}{c}{$\alpha=2$} & \multicolumn{2}{c}{$\alpha=5$} & \multicolumn{2}{c}{$\alpha=10$} \\
\cmidrule(lr){3-4} \cmidrule(lr){5-6} \cmidrule(lr){7-8} \cmidrule(lr){9-10}
 & & CA & RA & CA & RA & CA & RA & CA & RA \\
\midrule
\multirow{3}{*}{MNIST} & $1$ & 93.68 & 34.06 & 95.96 & 39.58 & 96.98 & 50.70 & 96.84 & 49.90 \\
& $2$ & 73.32 & 28.68 & 95.70 & 38.60 & 96.88 & 51.10 & 96.98 & 48.86 \\
& $5$ & 63.46 & 29.04 & 74.36 & 29.70 & 96.26 & 43.24 & \textbf{96.74} & \textbf{51.52} \\
\midrule
\multirow{3}{*}{FMNIST} & $1$ & 38.62 & 34.15 & 54.08 & 43.95 & 68.88 & 51.28 & 67.78 & 51.39 \\
& $2$ & 37.22 & 34.06 & 48.86 & 40.72 & 68.72 & 51.28 & \textbf{68.58} & \textbf{51.82} \\
& $5$ & 28.36 & 27.00 & 40.78 & 35.59 & 53.42 & 42.58 & 67.16 & 49.82 \\
\midrule
\multirow{3}{*}{CIFAR10} & $1$ & 33.40 & 23.22 & 39.84 & 24.85 & 41.00 & 25.30 & 40.96 & 24.95 \\
& $2$ & 33.46 & 23.50 & 39.34 & 24.52 & 42.72 & 25.68 & 40.68 & 24.79  \\
& $5$ & 18.94 & 14.57 & 35.12 & 22.78 & 41.16 & 25.08 & \textbf{43.10} & \textbf{25.38} \\
\bottomrule
\end{tabular}}
\end{table}

\begin{table*}[t]
\centering
    \caption{Clean accuracy and robust accuracy (i.e., FGSM, BIM, PGD-40, Square, and AA) comparison on MNIST, Fashion-MNIST, and CIFAR-10 under both IID and non-IID settings. The best results are in \textbf{bold} and second with \underline{underline}.}
    \label{tab:CA_RA_comparison}
\scalebox{0.97}{
\begin{tabular}{c|c|cccccc|cccccc}
\toprule
/\  & Setting & \multicolumn{6}{c|}{Non-IID} & \multicolumn{6}{c}{IID} \\
\midrule
Dataset & Methods & Clean & FGSM & BIM & PGD-40 & Square & AA & Clean & FGSM & BIM & PGD-40 & Square & AA \\
\midrule
\multirow{6}{*}{MNIST} 
 & FST & 91.38 & 31.08 & 25.28 & 0.50 & 0.58 & 0.00 & 85.74 & 29.46 & 21.78 & 1.44 & 0.10 & 0.00  \\
 & FedPGD & 72.96 & 39.30 & 47.72 & 19.98 & 8.68 & 4.44 & 60.24 & 26.08 & 30.12 & 13.20 & 8.04 & 6.08 \\
 & FedALP & 71.38 & 35.86 & 46.34 & 18.46 & 7.30 & 4.46 & 59.08 & 25.42 & 28.30 & 13.76 & 8.92 & \underline{7.82} \\
 & FedTRADES & 72.90 & 38.78 & 47.78 & 19.54 & 8.46 & 4.54 & 60.62 & 26.04 & 30.16 & 13.42 & 8.38 & 6.08  \\
 & FedALC & \underline{95.14} & \underline{64.04} & \underline{71.94} & \underline{39.52} & \underline{11.18} & \underline{8.04} & \underline{94.50} & \underline{59.92} & \underline{68.62} & \underline{36.38} & \underline{10.64} & 7.28  \\
  \cmidrule{2-14}
 & FatCC (ours) & \textbf{96.74} & \textbf{73.04} & \textbf{80.46} & \textbf{55.68} & \textbf{25.06} & \textbf{23.38} & \textbf{96.56} & \textbf{72.44} & \textbf{80.14} & \textbf{57.06} & \textbf{28.72} & \textbf{27.14} \\
 \midrule
 \multirow{6}{*}{Fashion-MNIST} 
 & FST & 59.74 & 28.56 & 13.00 & 12.62 & 12.66 & 11.14 & 58.88 & 33.66 & 15.54 & 15.16 & 15.44 & 14.40  \\
 & FedPGD & 37.72 & 25.72 & 22.96 & 22.90 & 20.72 & 20.02 & 41.58 & 28.10 & 25.78 & 25.48 & 20.00 & 19.62 \\
 & FedALP & 38.40 & 27.28 & 24.60 & 24.28 & 21.20 & 20.24 & 43.82 & 30.02 & 26.82 & 26.72 & 21.42 & 20.80 \\
 & FedTRADES & 37.78 & 25.48 & 22.78 & 22.44 & 20.40 & 19.54 & 41.18 & 28.06 & 25.58 & 25.10 & 19.74 & 19.42 \\
 & FedALC & \underline{65.46} & \underline{48.14} & \underline{43.88} & \underline{44.12} & \underline{38.50} & \underline{37.84} & \underline{63.64} & \underline{50.22} & \underline{46.96} & \underline{47.08} & \underline{39.64} & \underline{39.00}\\
\cmidrule{2-14}
 & FatCC (ours) & \textbf{68.58} & \textbf{57.10} & \textbf{54.10} & \textbf{54.14} & \textbf{46.92} & \textbf{46.82} & \textbf{71.98} & \textbf{60.60} & \textbf{56.36} & \textbf{56.40} & \textbf{48.06} & \textbf{47.46} \\
 \midrule
\multirow{6}{*}{CIFAR-10} 
 & FST    & 41.58 & 6.00  & 0.98  & 0.96  & 5.68  & 0.64 & 48.68 & 7.42 & 1.18 & 0.98 & 5.32 & 0.42 \\
 & FedPGD & 23.94 & 19.86 & 19.48 & 19.42 & 18.46 & 17.74 & 27.34 & 21.48 & 21.08 & 21.10 & 17.66 & 16.30 \\
 & FedALP & 23.80 & 19.38 & 18.88 & 18.88 & 18.42 & 17.86 & 27.38 & 21.68 & 21.03 & 21.06 & 17.70 & 16.28 \\
 & FedTRADES & 23.84 & 19.74 & 19.30 & 19.26 & 18.44 & 17.82 & 27.56 & 21.32 & 21.04 & 21.08 & 17.62 & 16.68 \\
 & FedALC & \underline{38.64} & \underline{26.38} & \underline{26.04} & \underline{25.56} & \underline{22.02} & \underline{20.13} & \underline{36.56} & \underline{26.18} & \underline{25.50} & \underline{25.44} & \underline{21.64} & \underline{20.16} \\
 \cmidrule{2-14}
 & FatCC (ours) & \textbf{43.10} & \textbf{28.64} & \textbf{27.22} & \textbf{27.20} & \textbf{23.04} & \textbf{20.78} & \textbf{45.54} & \textbf{30.14} & \textbf{28.46} & \textbf{28.36} & \textbf{22.90} & \textbf{20.52} \\
 \bottomrule
\end{tabular}}
\end{table*}

\textbf{Baselines.} To evaluate the robustness of our proposal and existing methods, we choose 5 different mainstream attack methods, including FGSM~\cite{goodfellow2014explaining}, BIM~\cite{kurakin2018adversarial}, PGD~\cite{madry2017towards}, Square~\cite{andriushchenko2020square}, and AA~\cite{croce2020reliable} attacks. In terms of adversarial defense methods, we integrate 3 well-known defense techniques: PGD~\cite{madry2017towards}, ALP~\cite{kannan2018adversarial}, and TRADES~\cite{zhang2019theoretically}, into FL framework and term them FedPGD, FedALP, and FedTRADES, respectively. Moreover, we compare FatCC with the other federated defense method FedALC~\cite{qiao2024noms_fedalc}. In addition, for a comprehensive comparison, we compare all methods with FST, where FST denotes the plain FL training strategy without the AT process. By default, all baselines are evaluated using 5 clients.

\textbf{Implementaion details.} 
Following previous work~\cite{qiao2023mp,mu2023fedproc,zizzo2020fat}, our work focuses on the typical label non-IID setting, where clients have different label distributions but the same feature distribution. This kind of label non-IID is usually simulated by Dirichlet distribution Dir($\gamma$)~\cite{yurochkin2019bayesian}, the smaller the value of $\gamma$ means the greater the skewness between clients, and vice versa. By default, we set the $\gamma$ to 0.5, and given that our goal is to evaluate the effectiveness of the proposed method, we randomly select 10\% samples for training from MNIST and Fashion-MNIST, while for the more complex task CIFAR-10, we randomly select 20\% samples for training. Following~\cite{goodfellow2014explaining}, we adopt the following AT settings: for MNIST, we set the perturbation bound to 0.3 and the step size to 0.01. For Fashion-MNIST, the perturbation bound is set to 32/255 with a step size of 8/255. For CIFAR-10, the perturbation bound is set to 8/255, and the step size is set to 2/255. In addition, we use the SGD optimizer and set local batch size, learning rate, and temperature as 128, 0.01, and 0.07, respectively. 

\subsection{Choosing $\alpha$ and $\beta$}
\label{subsec:choose}
As discussed in Section~\ref{subsec:local_re_weight}, the choice of $\alpha$ and $\beta$ has an impact on FatCC. $\beta$ controls the sensitivity to class frequency. A larger $\beta$ provides a greater difference between the majority and minority classes, while a smaller $\beta$ makes the response to class frequency flatter. A similar effect is introduced by $\alpha$, whose function is to control the strength of the overall weight. Given the distinct characteristics of each dataset, we explore the impact of various combinations of $\alpha$ and $\beta$ by heuristically selecting values from $\alpha \in \{1, 2, 5, 10\}$ and $\beta \in \{1, 2, 5\}$ for different datasets. We empirically choose the best trade-off between CA and RA, where CA refers to the averaged accuracy on clean images, while RA represents the averaged robust accuracy under 5 attacks, including FGSM, BIM, PGD-40, Square, and AA attacks. The results for MNIST, Fashion-MNIST and CIFAR-10 are reported in Table~\ref{tab:alpha_beta_chosen_all}. Empirically, we find that the best trade-off combination for MNIST and CIFAR-10 is $\alpha = 10$ and $\beta = 5$. For Fashion-MNIST, the trade-off combination is $\alpha = 10$ and $\beta = 2$.

\begin{figure*}[t]
\centering
\subfigure{\includegraphics[scale=0.31]{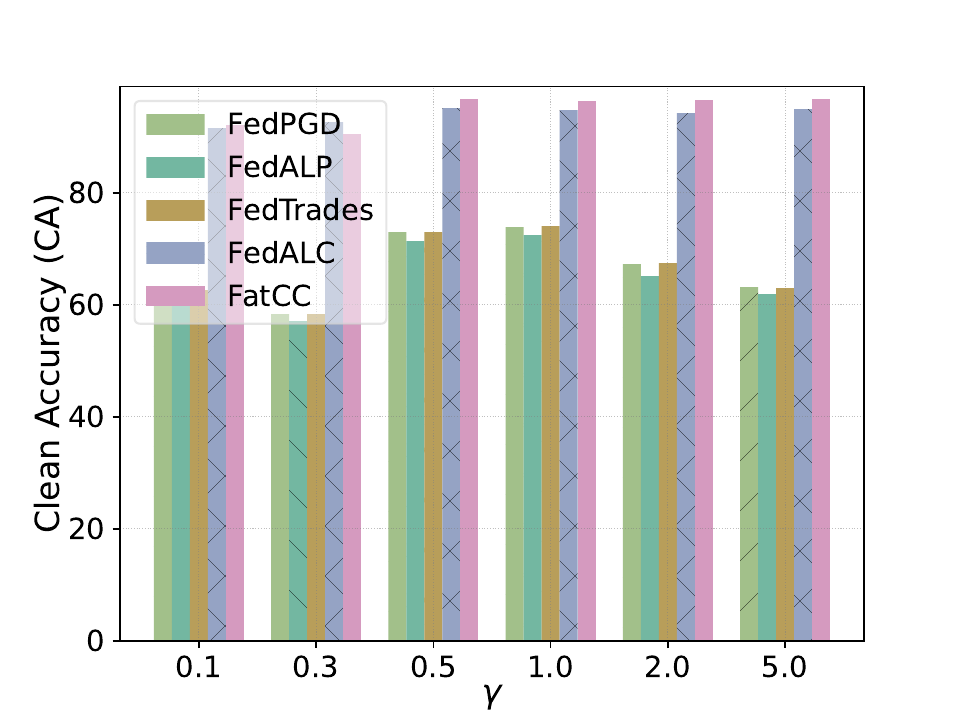}}
\subfigure{\includegraphics[scale=0.31]{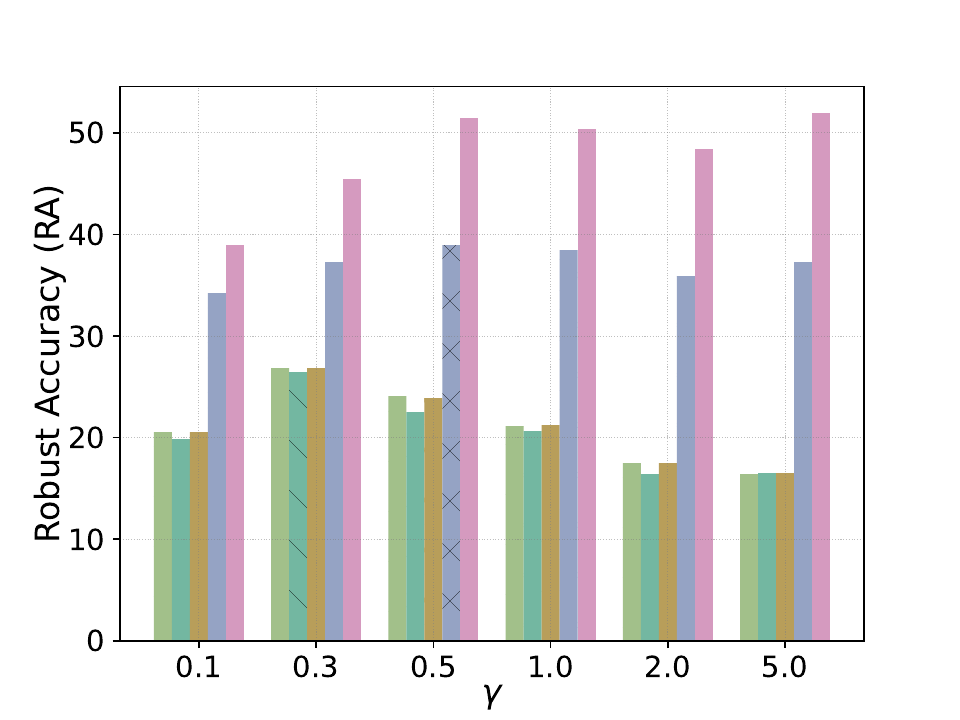}}
\subfigure{\includegraphics[scale=0.31]{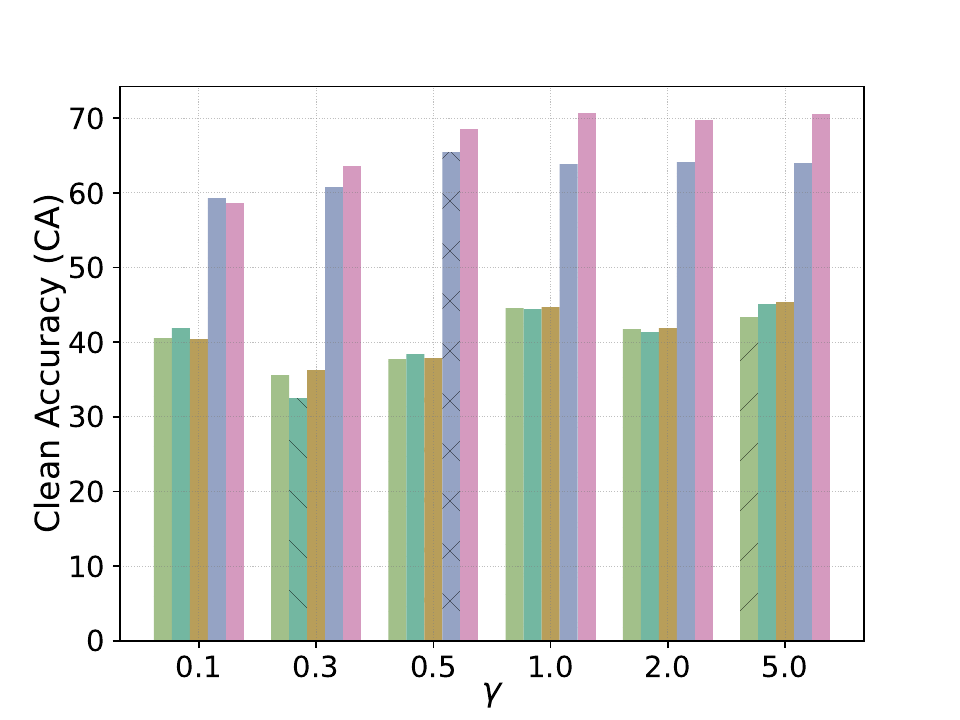}}
\subfigure{\includegraphics[scale=0.31]{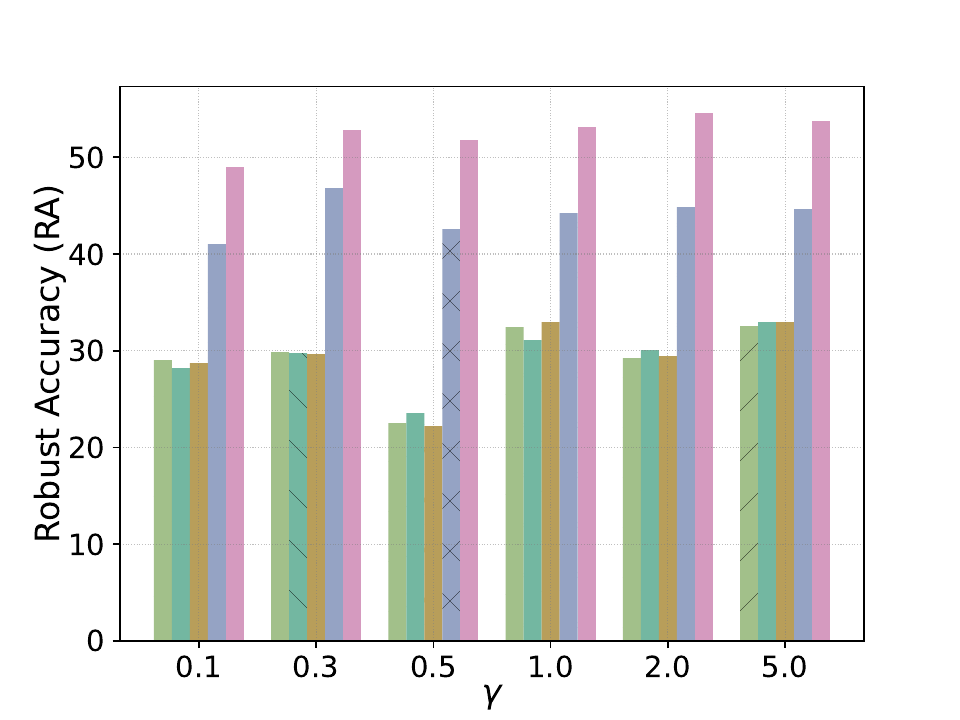}}
\caption{Illustration of CA and RA comparisons with varying levels of label skewness on MNIST and FashionMNIST datasets. The two figures on the left present comparisons under MNIST, while the two figures on the right depict comparisons under FashionMNIST.}
\label{fig:diff_gamma_nonIID}
\vspace{-5px}
\end{figure*}

\begin{figure*}[t]
\centering
\subfigure{\includegraphics[scale=0.31]{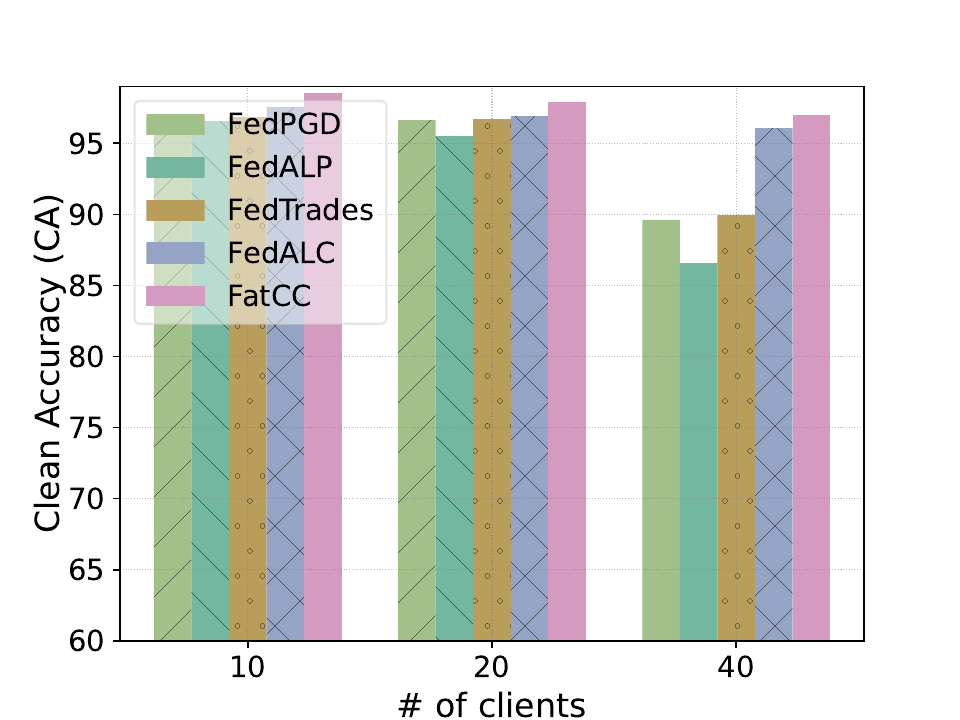}}
\subfigure{\includegraphics[scale=0.31]{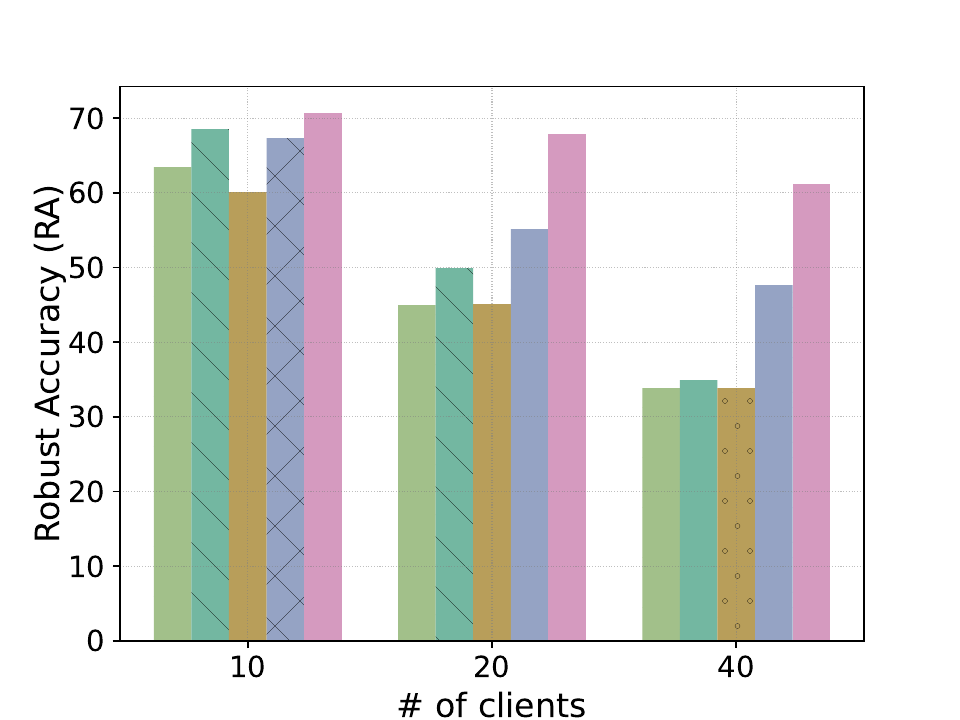}}
\subfigure{\includegraphics[scale=0.31]{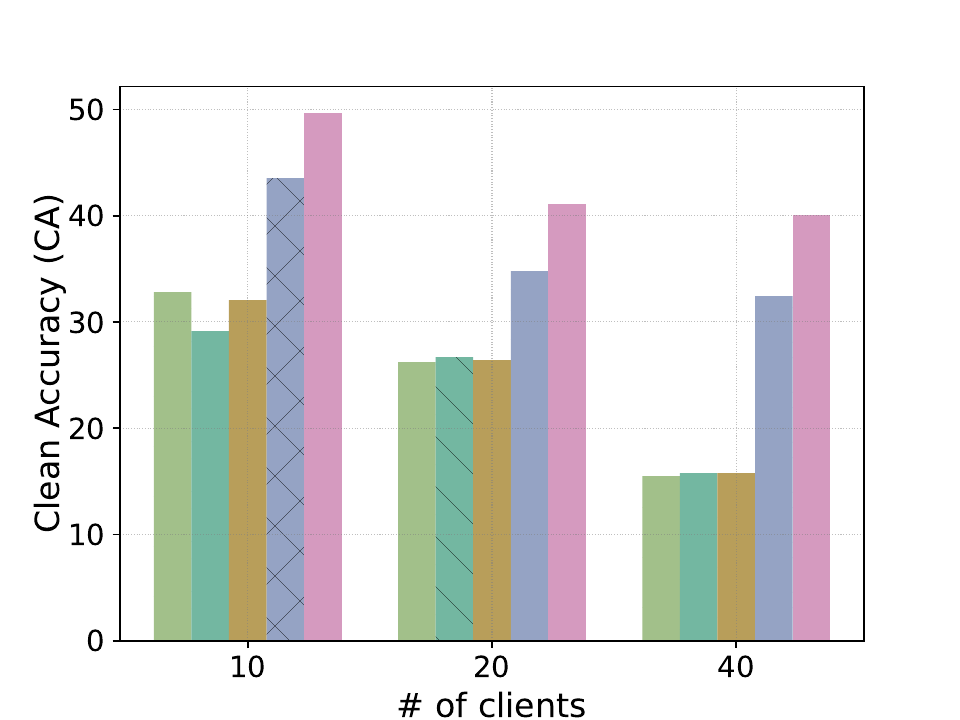}}
\subfigure{\includegraphics[scale=0.31]{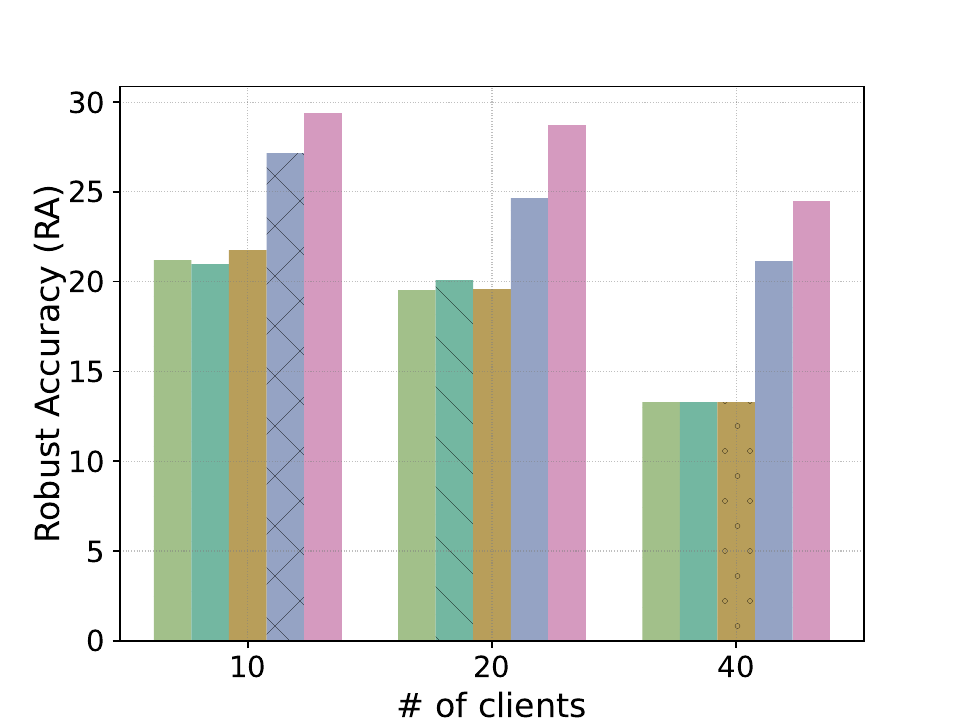}}
\caption{Illustration of CA and RA comparisons with different numbers of clients on MNIST and CIFAR-10 datasets with Dir(0.5). The two figures on the left present comparisons under MNIST, while the two figures on the right depict comparisons under CIFAR-10.}
\label{fig:diff_users_nonIID}
\vspace{-5px}
\end{figure*}

\subsection{Accuracy Comparison}
\label{subsec:accuracy_compar}
We implement FatCC and all baselines using Pytorch. We preliminarily 
compare CA and RA of all methods on clean images and AEs under non-IID and IID settings, respectively. The results are reported in Table~\ref{tab:CA_RA_comparison}, in which all the methods are calculated based on the average of the last 5 iterations. An overall trend can be observed that FatCC outperforms all baselines by a significantly large margin in terms of clean and robust accuracy.

Specifically, taking the results of Fashion-MNIST as an example, FatCC stands out as the top performer across all metrics. This includes both clean and robust accuracy, the former being measured under clean examples, while the latter being measured under various adversarial attacks such as FGSM, BIM, PGD-40, Square, and AA. Notably, compared with the second-best (i.e., FedALC), FatCC exhibits a 3.12\% increase in CA and a notable 9.32\% improvement in RA under the non-IID setting, where the RA value is calculated by the average of the above 5 attacks. Meanwhile, under the IID setting, FatCC significantly improves CA by 8.34\% and RA by 9.2\% compared with FedALC. A similar trend is also evident in MNIST and CIFAR-10, further highlighting the effectiveness of our proposed FatCC in not only enhancing RA but also maintaining a high level of CA. More notably, it is evident that the FST algorithm (i.e., without AT process) exhibits the lowest accuracy across all robustness comparison metrics compared to all other baselines. For example, in the case of CIFAR-10, the adversarial robust accuracy of FST is 0.64 and 0.42 under AA attack for non-IID and IID settings, respectively. This result further confirms our previous observation that adversarial attacks pose significant challenges to FL. Simultaneously, as indicated by the results in Table~\ref{tab:CA_RA_comparison}, it is noted that the method solely relying on standard AT (such as FedPGD) can somewhat improve adversarial accuracy; however, this improvement comes at the cost of a significant decrease in accuracy of clean samples. This is evident in CIFAR-10, where clean accuracy decreases from 41.58 for FST to 23.94 for FedPGD under the non-IID setting. This further confirms our previous observation that the straightforward adoption of the AT strategy to FL for enhancing adversarial robustness has limited effectiveness.

\begin{figure*}[t]
\centering
\subfigure{\includegraphics[scale=0.40]{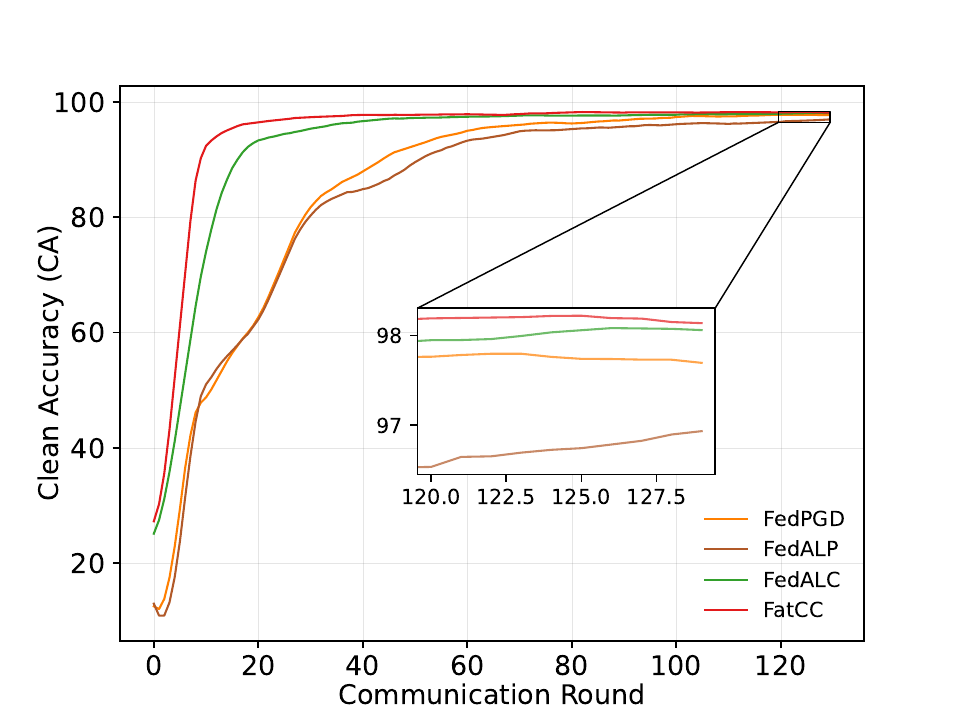}}
\subfigure{\includegraphics[scale=0.40]{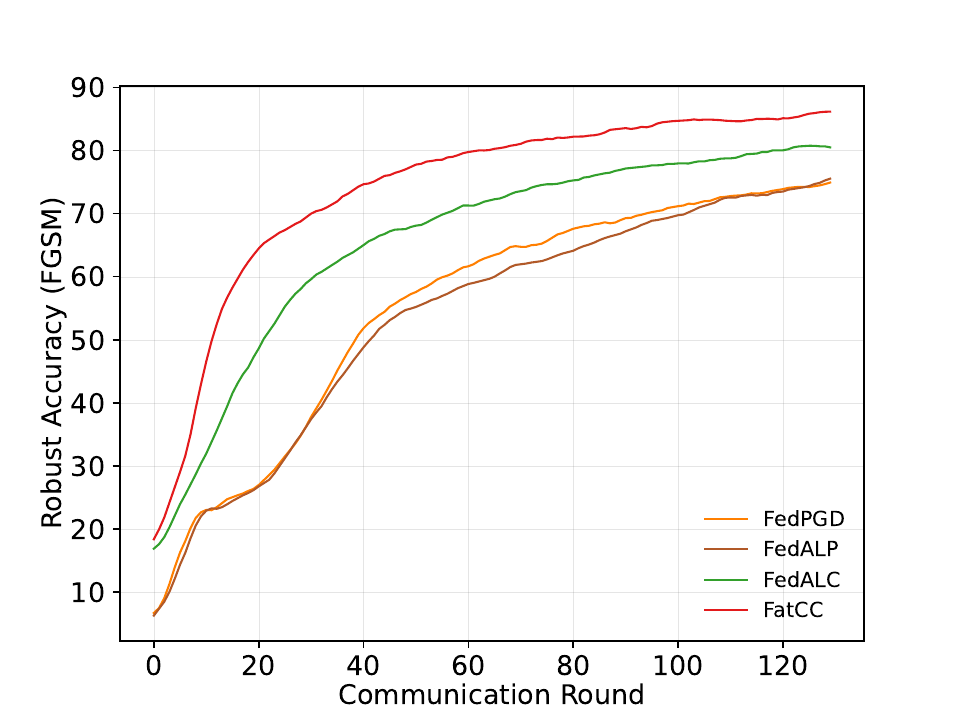}}
\subfigure{\includegraphics[scale=0.40]{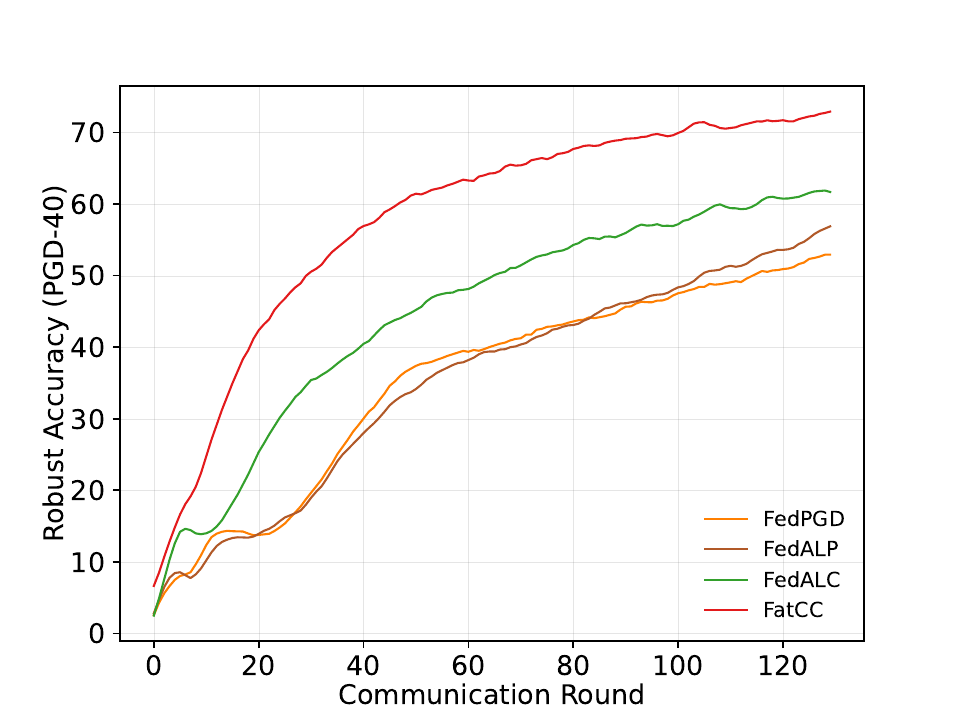}}
\caption{Comparison of communication efficiency of different benchmarks on CA, RA (FGSM), and RA (PGD-40) on MNIST. The comparisons start with CA, followed by RA under FGSM and PGD-40 attacks, respectively, from left to right.}
\label{fig:com_effici_mnist}
\vspace{-5px}
\end{figure*}

\begin{figure*}[t]
\centering
\subfigure{\includegraphics[scale=0.40]{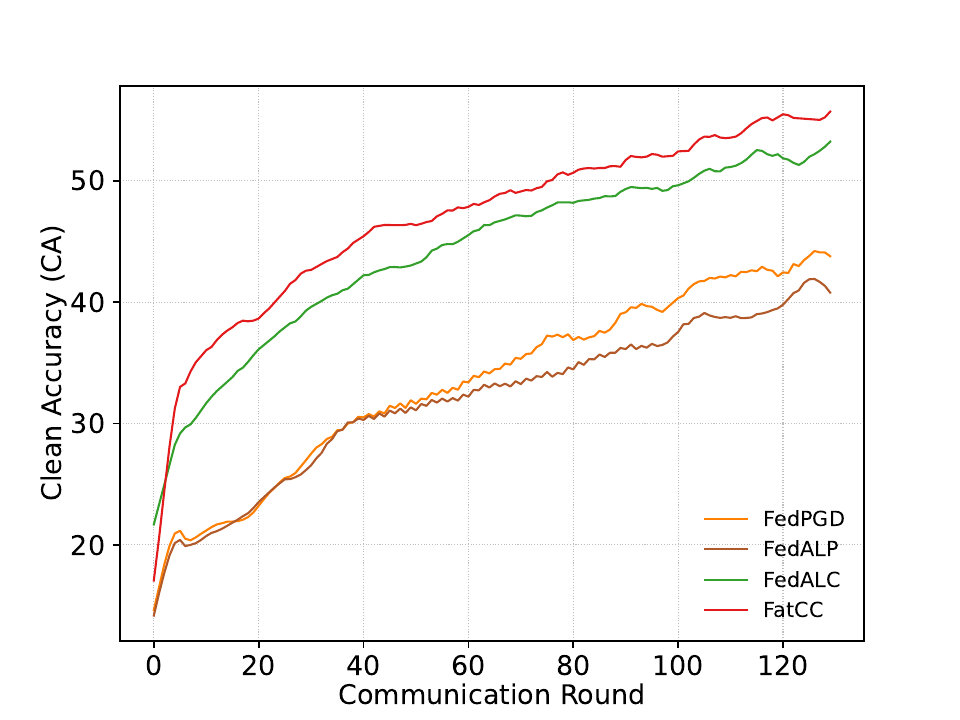}}
\subfigure{\includegraphics[scale=0.40]{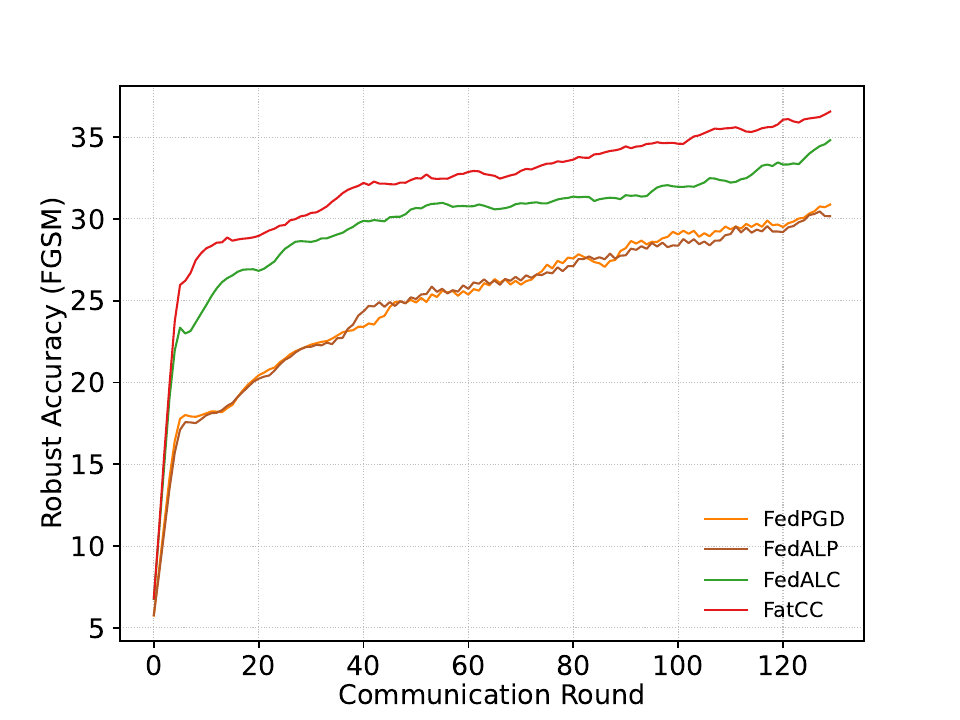}}
\subfigure{\includegraphics[scale=0.40]{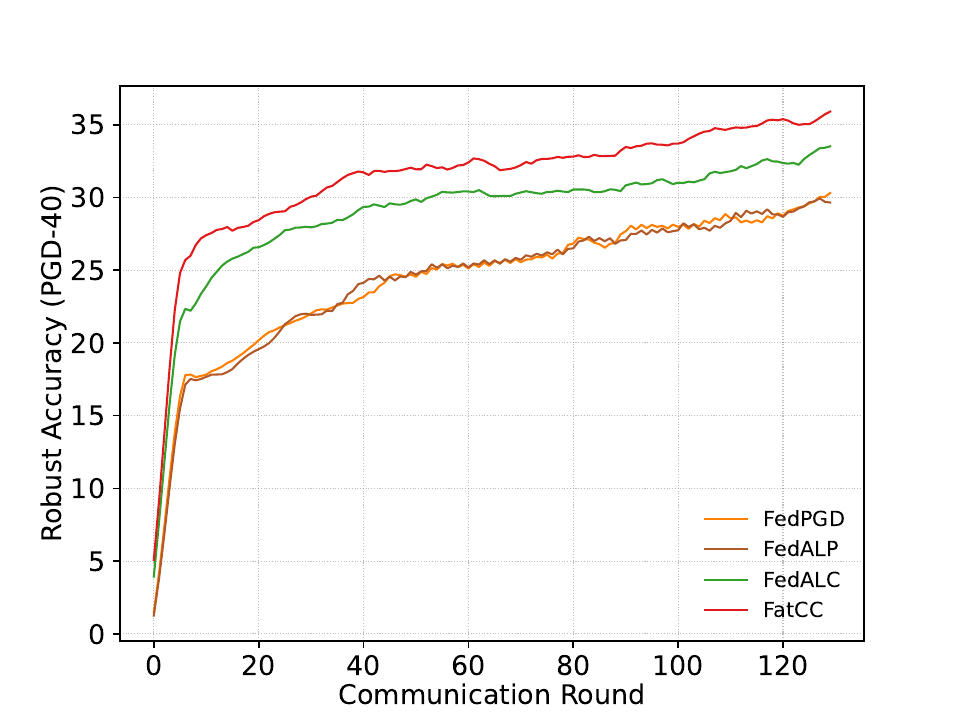}}
\caption{Comparison of communication efficiency of different benchmarks on CA, RA (FGSM), and RA (PGD-40) on CIFAR-10. The comparisons start with CA, followed by RA under FGSM and PGD-40 attacks, respectively, from left to right.}
\label{fig:com_effici_cifar10}
\vspace{-5px}
\end{figure*}

\subsection{Robustness Comparison}
\label{subsec:robust_compar}
\textbf{Different levels of non-IID.} As highlighted in the section above, the problem of non-IID data is considered a key challenge in federated adversarial environments. Meanwhile, given the challenges posed by varying levels of data heterogeneity that may exist in real-world scenarios, it becomes imperative to evaluate an algorithm that can demonstrate robustness across varying degrees of heterogeneity for real-world deployments. Therefore, as shown in Figure~\ref{fig:diff_gamma_nonIID}, we compare our method's CA and RA (the RA value is calculated by the average of FGSM, BIM, PGD-40, Square, and AA attacks) performance against various baselines under diverse levels of data heterogeneity. These levels span a broad range of heterogeneous coefficient gamma values, including 0.1, 0.3, 0.5, 1.0, 2.0, and 5.0. An overall observation reveals that, under both CA and RA metrics, FatCC consistently exhibits significant advantages over other baselines, with FatCC demonstrating particularly notable superiority in most cases. For example, with $\gamma$ set to 1.0 and Fashion-MNIST dataset is considered, it is observed that while FedALC already outperforms other baselines by approximately 19\% and 12\% in CA and RA, it is noteworthy that FatCC still surpasses FedALC by 7\% in CA and 9\% in RA. This demonstrates the effectiveness of our proposal in improving both CA and RA.

\textbf{Different numbers of clients.} To further evaluate the robustness of our proposed method, we also investigate its performance under varying numbers of participating clients. As the number of clients increases, the sample distribution per client decreases, which poses more significant challenges to the federated training process. To avoid the possible situation where no samples are assigned to a certain client, different from the previous sample adoption method, we include all samples in the training process in this robustness evaluation scenario. We report the average CA and RA performance as the number of clients increases from 10 to 40 in Figure~\ref{fig:diff_users_nonIID}. All methods in this scenario follow a Dirichlet distribution with parameter 0.5. Several observations can be made based on the results in the figure. First, as the number of clients increases, the value of CA and RA decreases for all methods, including ours, which proves our intuition that more clients pose a greater challenge to FL. Second, FatCC outperforms other benchmarks at different client numbers, and FatCC still outperforms FedALC in most cases. For example, in the case of the CIFAR-10 dataset with 10 clients, the CA and RA of FedALC consistently surpass other baselines such as FedPGD and FedALP by at least 10\% and 5\%, respectively, while FatCC still maintains performance advantages of 6\% and 2\% over FedALC in CA and RA, respectively. Considering the above findings, we conjecture that the reason why these baselines are unable to defend against adversarial attacks in federated adversarial environments is that these defense methods are not specifically designed for federated heterogeneous environments. This finding highlights the potential for further improvements in defenses against adversarial attacks in federated environments, and emphasizes the necessity for researchers to develop specialized defense mechanisms tailored to federated settings.

\begin{table}[t]
\centering
\caption{Ablation study on the efficacy of different modules in our proposed framework.}
\vspace{-2px}
\label{tab:ablation}
\resizebox{0.95\linewidth}{!}{
\begin{tabular}{c|cccc}
\toprule
Dataset& \multicolumn{2}{c}{MNIST} & \multicolumn{2}{c}{CIFAR10} \\
\midrule
Metric & CA & RA & CA & RA \\
\midrule
FedPGD (Base) & 72.96 & 24.02 & 23.94 & 18.99 \\
FatCC (w/o logit calibration) & 94.48 & 37.64 & 33.70 & 23.05  \\
FatCC (w/o feature contrast) & 95.60 & 41.81 & 35.40 & 24.13 \\
FatCC & \textbf{96.74} & \textbf{51.52} & \textbf{43.10} & \textbf{25.38}  \\
\bottomrule
\end{tabular}}
\vspace{-5px}
\end{table}

\subsection{Communication Efficiency Comparison}
\label{subsec:communi_compar}
The communication efficiency comparisons of different benchmarks based on MNIST and CIFAR-10 are shown in Figure~\ref{fig:com_effici_mnist} and Figure~\ref{fig:com_effici_cifar10}. We conduct experiments utilizing all samples for each dataset, setting the Dirichlet parameter to 1.0, and the number of clients for MNIST and CIFAR-10 is configured to be 20 and 10, respectively. Both sets of results demonstrate that our proposed method achieves not only higher accuracy but also faster convergence.

More specifically, in Figure~\ref{fig:com_effici_mnist}, FatCC exhibits an approximate 2\% improvement over the base FedPGD in CA. Similarly, in the more challenging CIFAR-10 task, the CA of FatCC exceeds that of FedALC and FedPGD by approximately 3\% and 11\%, respectively. For the comparison of adversarial robustness, we focus solely on illustrating the RA under FGSM and PGD-40 attacks, as these are among the most widely used attack methods. From the results of the PGD-40 attack, for instance, we observe that for CIFAR-10, FatCC exhibits improvements of approximately 5\% over FedPGD. Similarly, for MNIST, the enhancements are notably higher, with FatCC surpassing FedALC and FedPGD by around 11\% and 20\%, respectively. Importantly, we observe that within the same number of communication rounds, FatCC quickly achieves significant improvement in accuracy compared to other baselines, which, to some extent, indicates that our proposal can converge faster.

\subsection{Ablation Study}
\label{subsec:ablation}
To analyze the efficacy of modules in our proposed framework, we conduct ablation studies to evaluate the impact of each component on the overall performance. Table~\ref{tab:ablation} shows the ablation results and several key observations can be made. First, the lack of local calibration or global alignment based on feature contrast leads to performance degradation of CA and RA on various datasets, which highlights the importance of logit calibration and feature contrast. For example, when considering the MNIST dataset, disabling the calibration strategy causes the CA performance to drop from 96.88 to 94.48, while disabling the alignment strategy causes the CA performance to drop from 96.88 to 95.12. Second, either local calibration or feature contrast can significantly improve performance compared to the base (FedPGD). This shows that our method can gain benefits not only from local calibration but also from global alignment strategies. Third, combining logit calibration and feature contrast can lead to better overall performance, which, to some extent, supports our motivation of exploiting the combination of logit calibration and feature contrast for both CA and RA improvement in adversarial federated environments.

\section{Conclusion}
\label{sec:conclusion}
This paper explores the adversarial attack and non-IID challenges in FL environments. We have proposed the FatCC framework, which integrates local calibration and global alignment strategies into the FAT framework to tackle these two challenges. The first strategy alleviates local biases in achieving adversarial robustness, while the second provides an unbiased global signal to guide each local AT, thus further enhancing accuracy. The two strategies complement each other, with the goal of achieving robust FL on non-IID data. Our proposal is demonstrated effective through extensive experiments, showing improvements in both CA and RA across multiple datasets.

\bibliographystyle{ieeetr}
\bibliography{bib_global} 

\end{document}